\documentclass[12pt]{article}
\usepackage{graphicx}
\usepackage{natbib} %comment out if you do not have the package
\usepackage{url} % not crucial - just used below for the URL 
\usepackage{epsfig, epsf, graphicx, subfigure, color}
\usepackage{pstricks, pst-node, psfrag}
\usepackage{amssymb,amsmath}
\usepackage{verbatim,enumerate}
\usepackage{rotating, lscape}
\usepackage{setspace}
\usepackage{bbm}
\usepackage{natbib}
\usepackage{amsthm}
\usepackage{tikz}
\usepackage{etoolbox} % for \ifnumcomp
\usepackage{listofitems} % for \readlist to create arrays
\usepackage{url}
\usepackage{appendix}
\usepackage{graphicx}
\usepackage{booktabs}
\usepackage{algorithm}
\usepackage{algpseudocode}
\usepackage{color}
\RequirePackage[colorlinks,citecolor=blue,urlcolor=blue]{hyperref}
\graphicspath{ {./images/} }

\newcommand{\blind}{0}
\newcommand{\bfs}{\mathbf{s}}
\begin{document}

\def\spacingset#1{\renewcommand{\baselinestretch}%
{#1}\small\normalsize} \spacingset{1}

%%%%%%%%%%%%%%%%%%%%%%%%%%%%%%%%%%%%%%%%%%%%%%%%%%%%%%%%%%%%%%%%%%%%%%%%%%%%%%

\if0\blind
{
  \title{\bf Bivariate DeepKriging for Large-scale
Spatial Interpolation of Wind Fields}
  \author{Pratik Nag\thanks{
    The authors gratefully acknowledge KAUST for funding this research. }\hspace{.2cm}\\
    Ying Sun \\
    CEMSE Division, Statistics Program, \\
    King Abdullah University of Science and Technology,Saudi Arabia.\\
    and \\
    Brian J Reich \\
    Department of Statistics, \\
    North Carolina State University, Raleigh, USA.}
    \date{}
  \maketitle
} \fi

\if1\blind
{
  \bigskip
  \bigskip
  % \bigskip
  \begin{center}
    {\LARGE\bf Title}
\end{center}
  \medskip
} \fi

\bigskip
\vspace{-15mm}
\begin{abstract}
High spatial resolution wind data play a crucial role in various fields such as climate, oceanography, and meteorology. However, spatial interpolation or downscaling of bivariate wind fields, characterized by velocity in two dimensions, poses a challenge due to their non-Gaussian nature, high spatial variability, and heterogeneity. While cokriging is commonly employed in spatial statistics for predicting bivariate spatial fields, it is suboptimal for non-Gaussian processes and computationally prohibitive for large datasets. In this paper, we introduce bivariate DeepKriging, a novel method utilizing a spatially dependent deep neural network (DNN) with an embedding layer constructed by spatial radial basis functions for predicting bivariate spatial data. Additionally, we devise a distribution-free uncertainty quantification technique based on bootstrap and ensemble DNN. We establish the theoretical basis for bivariate DeepKriging by linking it with the Linear Model of Coregionalization (LMC). Our proposed approach surpasses traditional cokriging predictors, including those utilizing commonly used covariance functions like the linear model of co-regionalization and parsimonious bivariate Mat\'ern covariance. We demonstrate the computational efficiency and scalability of the proposed DNN model, achieving computation speeds approximately 20 times faster than conventional techniques. Furthermore, we apply the bivariate DeepKriging method to wind data across the Middle East region at 506,771 locations, showcasing superior prediction performance over cokriging predictors while significantly reducing computation time.
\end{abstract}

\vspace{-3mm}
\noindent%
{\it Keywords:}  Bivariate spatial modelling, Deep learning, Feature embedding, Probabilistic prediction, Radial basis function, Spatial regression 
\vfill

\newpage
\spacingset{2} % DON'T change the spacing!
\section{Introduction}\label{sec:intro}

Cokriging \citep{goovaerts1998ordinary}, a multivariate extension of univariate kriging, is widely used for analysis and prediction of multivariate spatial fields in multivariate spatial statistics. Prediction (with uncertainty quantification) at new unobserved sites using the information gained from the observed locations is one of the common objectives of this strategy. For Gaussian random fields with known covariance structure, cokriging is the best linear unbiased predictor. However, in reality, this requirement of Gaussianity and properly stated covariance is rarely met. Because of this, modeling heavy-tailed or skewed data with complicated covariance structure requires a more adaptable prediction methodology.

Various approaches have been proposed to model non-Gaussian spatial data such as scale mixing Gaussian random fields \citep{fonseca2011non}, multiple indicator kriging \citep{journal1989non}, skew-Gaussian processes \citep{zhang2010spatial}, copula-based multiple indicator kriging \citep{agarwal2021copula} and Bayesian nonparametrics \citep{reich2015spatial}. To address the nonstationary behaviour of the random fields over a large region, non-stationary Mat\'ern covariance models \citep{nychka2002multiresolution,paciorek2003nonstationary,cressie1999classes,fuentes2001high} have been introduced. Trans-Gaussian random fields \citep{cressie1993noel,de2002bayesian} find nonlinear transformations which enables application of Gaussian processes on the transformed data. However, marginal transformation to normality may not confirm joint normality at multiple locations and may change important properties of the variable \citep{changyong2014log}. One common drawback of many of the approaches is that there is no straight forward implementation of these methods in the bivariate setting and also they may not be optimal for more general spatial datasets. Another issue is that most of these models are based on Gaussian processes which uses cholesky decomposition of the $N \times N$ covariance matrix which requires $\mathcal{O}(N^3)$ time and $\mathcal{O}(N^2)$ memory complexity where $N$ is the number of spatial locations. 

Recently, Deep neural network (DNN) based algorithms have proved to be the most powerful prediction methodologies in the field of computer vision and natural language processing \citep{lecun2015deep}. DNNs can handle more complex functions and in theory they can be applied to approximate any function which is appealing for modelling large-scale nonstationary and non-Gaussian spatial processes. DNNs are also computationally efficient and thus can be applied to large datasets. The computation time for training can be massively reduced by using GPUs \citep{najafabadi2015deep}. Due to this broad applicability of the neural networks, several researchers are trying to incorporate DNNs in the spatial problems \citep{wikle2022statistical}. \cite{cracknell2014geological} included spatial coordinates as features for DNNs. \cite{wang2019nearest} proposed a nearest neighbour neural network approach for Geostatistical modelling. \cite{zammit2021deep} used neural networks to estimate the warping function which transforms the spatial domain to fit stationary and isotropic covariance structure.  Convolutional neural networks (CNNs) \citep{krizhevsky2012imagenet} can capture the spatial dependence successfully, but require a huge amount of data on regular grid for model training. However, in environmental statistics scenarios one of the main objective is to give spatial interpolation at unobserved locations for irregular grid datasets which is infeasible for CNN modelling. Most of these DNN based methods are developed for univariate data and only concentrate on point predictions and ignore the prediction interval estimations. Recently, there has been few works which proposes density function estimation of the predictive distribution using neural networks, for example \cite{li2019deep} proposed a discretized density function approach and predict the predictive distribution probabilities by training a neural network classifier. \cite{neal2012bayesian} and \cite{posch2019variational} applied Bayesian inference methodologies to neural networks to predict uncertainties via posterior distributions. But none of these methods are directly applicable to spatial applications.

To address these drawbacks \cite{chen2020deepkriging} introduced a spatially dependent deep neural network structure called DeepKriging, for univariate spatial prediction. They used basis functions to capture spatial dependence and use them as features to fit the model. They also provided an approach for uncertainty quantification by an histogram based approach. However, the histograms require thresholds which are mostly deterministic and choice of the thresholds effects the results drastically.  

The objective of this study is to establish a nonparametric statistical framework for conducting bivariate spatial prediction and estimating prediction uncertainties. Our proposed approach, a bivariate extension of DeepKriging (BDK), aims to address these limitations. We introduce a spatially dependent neural network by incorporating an embedding layer of spatial coordinates using basis functions. Additionally, we propose a bootstrap and ensemble neural network-based prediction interval for prediction uncertainty. BDK is non-parametric, making it more versatile and applicable to non-Gaussian, nonstationary, and even categorical prediction problems. Through simulations, we demonstrate that this approach yields comparable results to cokriging when the underlying process is Gaussian, and it surpasses traditional statistical methodologies in non-Gaussian and nonstationary scenarios. Moreover, our proposed approach significantly reduces computation time compared to traditional methods.

While architecturally BDK is a straightforward extension of univariate DeepKriging, integrating this methodology with traditional statistical methods remains a challenging task. In this paper, we illustrate that under certain conditions, the linear model of coregionalization (LMC) represents a special case of BDK. Additionally, we introduce a novel methodology for computing prediction intervals using bootstrap and ensemble neural networks. Additionally, the execution of BDK is computationally twenty times faster than traditional statistical approaches, effectively overcoming the bottleneck associated with applying this method to large datasets. 

The remainder of our paper is structured as follows. Section~\ref{sec:method} outlines the proposed methodology. Section~\ref{sec:simulation} compares the performance of the proposed method with traditional approaches. Lastly, Section~\ref{sec:application} applies the BDK method to Saudi Arabian wind data. More detailed information on the exploratory data analysis of the wind dataset and further results on simulation studies can be found in the Supplementary materials.

\section{Bivariate DeepKriging }\label{sec:method}

\subsection{Background}

    Let $\{\mathbf{Y(\bfs)}=\{Y_1(\bfs),Y_2(\bfs)\}^T, s \in D\}$, $D \subseteq \mathbb{R}^{p}$, be a bivariate spatial process, and $\{\mathbf{Z}(\bfs_1), \\ \mathbf{Z}(\bfs_2),  \ldots ,\mathbf{Z}(\bfs_N)\}$ be the realization of the process at $N$ spatial locations, where $\mathbf{Z}(\bfs) = \{Z_1(\bfs),Z_2(\bfs)\}^T$.

A classical spatial model assumes 
\begin{equation}
    \mathbf{Z}(\bfs) = \mathbf{Y}(\bfs) + \boldsymbol{\epsilon}(\bfs)
    \label{eq:obs}
\end{equation}
where $\boldsymbol{\epsilon}(\bfs) = \{\epsilon_1(\bfs),\epsilon_2(\bfs)\}^T$, known as the nugget effect, is a bivariate white noise process with variances $\sigma_1^2(\bfs)$ and $\sigma_2^2(\bfs)$, respectively.

Given observations $\mathbf{Z}(\bfs)$, one of the main objectives of spatial prediction is to find the best predictor of the true process $\mathbf{Y}(\bfs_0)$ at the unobserved location $\bfs_0$. The optimal predictor with parameter set $\boldsymbol{\theta}$ can be defined by minimizing the expected value of the loss function, that is,
$$
\hat{\mathbf{Y}}^{opt}(\bfs_0|\mathbf{Z}_N) = \mathbb{E}\{L(\hat{\mathbf{Y}}(\bfs_0),\mathbf{Y}(\bfs_0))| \mathbf{Z}_{N}\} = \operatorname*{argmin}_{\hat{\mathbf{Y}}} R(\hat{\mathbf{Y}}(\bfs_0)| \mathbf{Z}_N),
$$
where $\hat{\mathbf{Y}}^{opt}(\bfs_0|\mathbf{Z}_N)$ is the optimal predictor given $\mathbf{Z}_{N} = \{\mathbf{Z}(\bfs_1),\mathbf{Z}(\bfs_2), \ldots ,\mathbf{Z}(\bfs_N)\}$. The function $R(\cdot)$ is the risk function, which is the approximation of the loss $L$. 

Bivariate kriging, also known as Cokriging, is the best linear unbiased predictor \citep{stein1991universal,goovaerts1998ordinary} for bivariate spatial prediction. The spatial process $\mathbf{Y(s)}$ is typically modeled as
$$\mathbf{Y}(\bfs) = \boldsymbol{\mu}(\bfs) + \boldsymbol{\gamma}(\bfs), $$
where $\boldsymbol{\gamma}(\bfs)=\{\gamma_1(\bfs),\gamma_2(\bfs)\}$ is a spatially-dependent zero-mean bivariate random process with cross-covariance function $\mathbf{C}_{u,v}(\bfs_i,\bfs_j)= \text{cov}(\gamma_u(\bfs_i),\gamma_v(\bfs_j))$, $u,v = 1,2$. The mean structure $\boldsymbol{\mu}(\bfs)$ can be modeled as $\boldsymbol{\mu}(\bfs) = \mathbf{X}(\bfs)^T\boldsymbol{\beta}$, where 
\[
\mathbf{X}(\bfs) = \begin{bmatrix} \mathbf{X}_1(\bfs)^T & \boldsymbol{0}\\
\boldsymbol{0} & \mathbf{X}_2(\bfs)^T
\end{bmatrix}
\]
and $\boldsymbol{\beta} = \{\boldsymbol{\beta}_1 , \boldsymbol{\beta}_2 \}^T$ is a $(p_1 + p_2) \times 1$ vector of coefficients corresponding to the covariates. Let $\mathbf{Z}_{vec} = \{Z_1(\bfs_1),\ldots,Z_1(\bfs_N), Z_2(\bfs_1), \ldots,Z_2(\bfs_N)\}^T$ be a $2N \times 1$ vector, $\mathbf{C}$ is the cross-covariance matrix of order $2N \times 2N$, and 
\[
\mathbf{X} = \begin{bmatrix} \{\mathbf{X}_1(\bfs_1)^T,...,\mathbf{X}_1(\bfs_N)^T\}^T & \boldsymbol{0}\\
\boldsymbol{0} & \{\mathbf{X}_2(\bfs_1)^T,...,\mathbf{X}_2(\bfs_N)^T\}^T
\end{bmatrix}
\]
is a $2N \times (p_1 + p_2)$ matrix of covariates. Then with $\mathbf{V} = (\mathbf{X}^T\mathbf{C}^{-1}\mathbf{X})^{-1}$, the kriging prediction at an unobserved location $\bfs_0$ is 
$$\hat{\mathbf{Y}}_{vec}(\bfs_0) = \mathbf{X_0}^T\hat{\boldsymbol{\beta}} + \mathbf{C_0}^T\mathbf{C}^{-1}(\mathbf{Z}_{vec}-\mathbf{X}\hat{\boldsymbol{\beta}})$$
where $\hat{\boldsymbol{\beta}} = \mathbf{V}\mathbf{X}^T\mathbf{C}^{-1}\mathbf{Z}_{vec}$, the generalized least square estimator of $\boldsymbol{\beta}$, $\mathbf{X_0}=\mathbf{X}(\bfs_0)$ is the $2 \times (p_1 + p_2)$ matrix of covariates at location $\bfs_0$, $\mathbf{C_0}$ is the $2N \times 1$ cross-covariance matrix between the response at location $\bfs_0$ and the observed locations.

The choice of cross-covariance functions in cokriging prediction plays a vital role in capturing the spatial dependence among locations and also among variables for multivariate data.

Among existing models, the multivariate Mat\'ern cross-covariance function \citep{apanasovich2012valid} is one of the most popular classes of Mat\'ern covariance function. The parsimonious Mat\'ern covariance function can be defined as
\begin{equation}
\begin{aligned}
\mathbf{C}_{u,v}(\mathbf{h}) = \sigma_{uv} M\left(\mathbf{h}\middle|\nu_{uv},\sqrt{\alpha_{uv}^2}\right), \quad \mathbf{h} = \mathbf{d}(\bfs_i,\bfs_j)  \in \mathbb{R},
\label{eq:2}
\end{aligned}
\end{equation}
where $\mathbf{d}(\bfs_i,\bfs_j) = \| \bfs_i - \bfs_j \|$ is the Euclidean distance between the locations, $M$ is the Mat\'ern correlation function \citep{cressie1999classes}, $\nu$ is the smoothness, and $\alpha$ is the range parameter. Constraints on these parameters have been derived in \cite{apanasovich2012valid}.

Another example of a cross-covariance function is the linear model of coregionalization (LMC); \citep{genton2015cross}. In this approach, a multivariate random field is represented as a linear combination of $r$ independent random fields with potentially different spatial correlation functions. For $r=2$, we can write the process $\boldsymbol{\gamma}(s)$ in terms of the latent processes as $\boldsymbol{\gamma}(\bfs) = \mathbf{AU(\bfs)}$, where $\mathbf{A}$ is the $2 \times 2$ matrix of weights and $\mathbf{U(s)} = \{U_1(\bfs),U_2(\bfs)\}^T$, $\text{cov}(U_1,U_2) = 0$. Following this architecture, the LMC covariance function of a bivariate random field can be written as  
\begin{equation}
\begin{aligned}
\mathbf{C}_{ij}(\mathbf{h})= \sum_{k=1}^{r}\rho_k(h)A_{ik}A_{jk}, \quad \mathbf{h} \in \mathbb{R}^d, \quad 1 \leq r \leq 2,
\label{eq:4}
\end{aligned}
\end{equation}
with $\rho_k(\cdot)$ as valid stationary correlation functions of $U_k(\cdot)$, and $A = (A_{ij})_{i,j=1}^{2,r}$ is a $2\times r$ full-rank matrix. 

Although cokriging is the best linear unbiased predictor, it suffers from several limitations. First, maximum likelihood estimation (MLE) for Gaussian processes requires Cholesky factorization of the covariance matrix, which is computationally expensive, making it infeasible to implement traditional cokriging for large datasets. Moreover, many real-life datasets are non-Gaussian, and cokriging is not an optimal predictor in those scenarios.

\subsection{DeepKriging for bivariate spatial data} \label{sec:deepkriging}

    DeepKriging, proposed by \cite{chen2020deepkriging}, addresses these issues for univariate processes by rephrasing the problem as regression and introducing a spatially dependent neural network structure for spatial prediction. They have used the radial basis functions to embed the spatial locations into a vector of weights and passed them to the DNNs as inputs allowing them to capture the spatial dependence of the process. 
    Similar to \cite{chen2020deepkriging}, we propose to use spatial basis functions for the bivariate processes to construct the embedding layer instead of passing the coordinates of the observations directly to the NN. This idea is motivated by the multivariate Karhunen-Lo\'eve expansion \citep{daw2022overview} of a multivariate spatial random field, i.e., a bivariate spatial process $\boldsymbol{\gamma}(\bfs)$ can be represented as  
    $$
    \boldsymbol{\gamma}(\bfs) = \sum_{b=1}^{\infty} \{ w_{b,1}\phi_{b,1}(\bfs),w_{b,2}\phi_{b,2}(\bfs) \}^T
    $$
    where $w_{b,u}$'s are independent random variables and $\phi_{b,u}(\bfs)$'s are the pairwise orthonormal basis functions corresponding to variable $u, u =1,2$. Therefore, a bivariate process can be approximated by
    $$
    \boldsymbol{\gamma}(\bfs) \approx \sum_{b=1}^{K} \{ w_{b,1}\phi_{b,1}(\bfs),w_{b,2}\phi_{b,2}(\bfs) \}^T,
    $$
    where the weights $w_{b,u}$ can be estimated by minimizing the risk function $R$. Hence the spatial problem can be viewed as multioutput linear regression \citep{borchani2015survey} with these basis functions $\phi_{b,u}$ as covariates.
    
    Among many existing basis functions, such as spline basis functions \citep{wahba1990spline}, wavelet basis functions \citep{vidakovic2009statistical}, and radial basis functions \citep{hastie2001elements}, we have chosen the multi-resolution compactly supported Wendland radial basis function \citep{nychka2015multiresolution} defined via $w(d) = \frac{(1-d)^6}{3}(35d^2 + 18d + 3)\mathbf{1}\{0\leq d \leq 1 \}$. The spatial basis functions are then defined as $\phi_i(\bfs) = w(\left\|\bfs-\mathbf{u}_i\right\|/\theta)$ with $\theta$ as the bandwidth parameter and anchor points (spatial locations) $\{\mathbf{u}_1,\mathbf{u}_2,...,\mathbf{u}_{K} \}$. We take $\{\mathbf{u}_1,\mathbf{u}_2,...,\mathbf{u}_{K} \}$ to be a square grid of locations covering the spatial domain and $\theta$ to be 2.5 times the distance between adjacent anchor points. By selecting various $K$ values, one can produce distinct sets of basis functions. This will enable us to record both the long- and short-range spatial dependencies. We must combine these basis functions into a single vector in order to send them to the neural network.
    See \cite{chen2020deepkriging} for detailed discussions on the basis function embedding. Therefore, these co-ordinates embedding can now be used as an input to the DNNs along with covariates $\mathbf{X}_{vec}(\bfs)=\{\mathbf{X}_1(\bfs)^T,\mathbf{X}_2(\bfs)^T\}^T$ which will better capture the spatial features of the data. Note that under our assumption of basis functions, $\boldsymbol{\phi}_{.,1} = \boldsymbol{\phi}_{.,2}$ for location $\bfs$, hence we can reduce our basis function set to $\boldsymbol{\phi}(\bfs) = \{ \phi_{1}(\bfs),...,\phi_{K}(\bfs)\}^T$ where $\phi_{b}(\bfs) = \phi_{b,1}(\bfs), \ b = 1,2,...,K$. 
    
    We have used a multi-output deep neural network structure to build the bivariate DeepKriging framework. We define $\mathbf{X}_{\phi}(\bfs)$ as the vector of inputs containing the embedded vector of basis functions $\boldsymbol{\phi}(\bfs)$ and the covariates $\mathbf{X}_{vec}(\bfs)$. Then $\mathbf{X}_{\phi}(\bfs) = (\boldsymbol{\phi}(\bfs)^T,\mathbf{X}_{vec}(\bfs)^T)^T$. Taking $\mathbf{X}_{\phi}(\bfs)$ as the inputs to a neural network with L layers and $M_l$ nodes in layer $l=1,...,L$ DNN can be specified as, 
        \begin{equation}
        \begin{aligned}
            \mathbf{h}_1(\mathbf{X}_{\phi}(\bfs)) &= \mathbf{W}_1 \mathbf{X}_{\phi}(\bfs) + \mathbf{b}_1, \ \mathbf{a}_1(\bfs)=\psi(\mathbf{h}_1(\mathbf{X}_{\phi}(\bfs))) \\
            \mathbf{h}_2(\mathbf{a}_1(\bfs)) &= \mathbf{W}_2 \mathbf{a}_1(\bfs) + \mathbf{b}_2, \ \mathbf{a}_2(\bfs)=\psi(\mathbf{h}_2(\mathbf{a}_1(\bfs))) \\
            . . &. \\
            \mathbf{h}_L(\mathbf{a}_{L-1}(\bfs)) &= \mathbf{W}_L \mathbf{a}_{L-1}(\bfs) + \mathbf{b}_L, \ \mathbf{f}_{NN}(\mathbf{X}_{\phi}(\bfs))=\mathbf{h}_L(\mathbf{a}_{L-1}(\bfs)), \\
        \label{eq:6}
        \end{aligned}
        \end{equation}
        where $\mathbf{f}_{NN}(\mathbf{X}_{\phi}(\bfs)) = \{f_{NN_1}(\mathbf{X}_{\phi}(\bfs)), f_{NN_2}(\mathbf{X}_{\phi}(\bfs))\}^T$ is the prediction output from the model, the weight matrix $\mathbf{W}_l$ is a $M_l \times M_{l-1}$ matrix of parameters and $\mathbf{b}_l$, a $M_l \times 1$ vector is the bias at layer $l$. The parameter set of this network is  $\boldsymbol{\theta} = \{ \mathbf{W}_l, \mathbf{b}_l : l = 1,2,...,L\}$. $\psi(\cdot)$ signifies the activation function which in our case is the rectified linear unit ReLU \citep{schmidt2020nonparametric}. The final layer does not contain any activation function and provides just the linear output allowing it to take any value in $\mathbb{R}^2$.
        
\begin{figure}[h]
    \centering
   \includegraphics[scale=0.43]{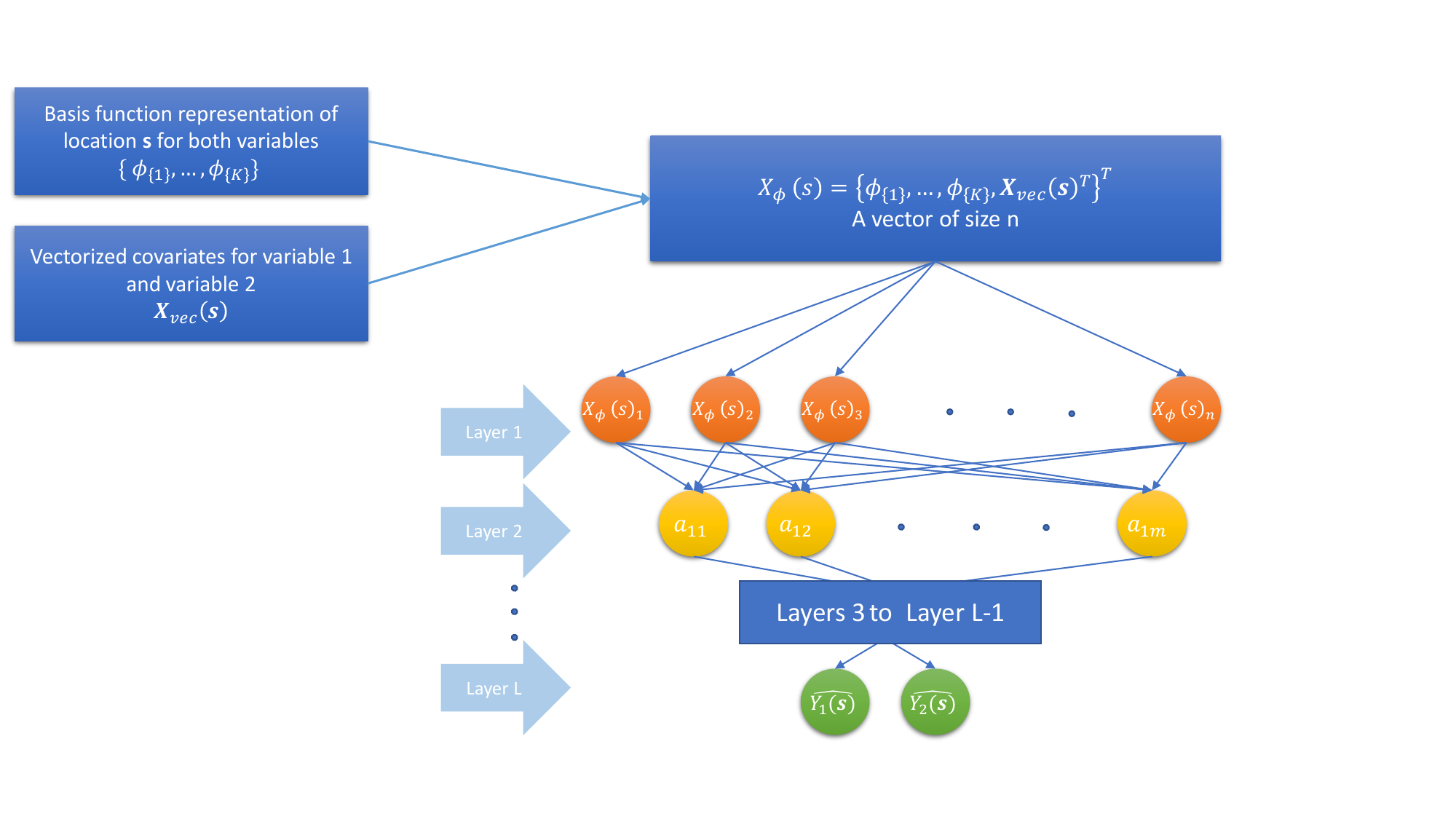}
   \caption{Structure of the spatially dependent neural network with embedding}
    \label{fig:NN_structure}
\end{figure}

        Figure \ref{fig:NN_structure} gives a graphical illustration of the spatially dependent neural network structure.

    The Bivariate DeepKriging (BDK) is a multivariate extension of the univariate DeepKriging. Here we propose to approximate the optimal predictor $\hat{Y}^{opt}(\bfs_0)$ with the help of deep neural network structure, i,e; $\hat{Y}^{opt}(\bfs_0) \approx \mathbf{f}_{NN}(\mathbf{X}_{\phi}(\bfs_0))$  We define the optimal neural network based predictor as $\mathbf{f}_{NN}^{opt}(\mathbf{X}_{\phi}(\bfs_0)) = \operatorname*{argmin}_{\mathbf{f}_{NN}} R(\mathbf{f}_{NN}(\mathbf{X}_{\phi}(\bfs_0))| \mathbf{Z}_N)$. where $\mathbf{f}_{NN} \in \mathcal{F}$ is a multioutput function in the domain of functions $\mathcal{F}$ which are feasible by the defined neural network structure. Note that here $\mathbf{f}_{NN}$ is conditioned on the parameter $\boldsymbol{\theta}$. Hence, one can estimate $\mathbf{f}_{NN}^{opt}(\cdot)$ by minimizing the empirical version of the risk function $R(\cdot)$ based on $\boldsymbol{\theta}$ given as
    \begin{equation}
    R\{\mathbf{f}_{NN}(\mathbf{X}_{\phi}(\bfs))| \boldsymbol{\theta},\mathbf{Z}_N\} = \frac{1}{N}\sum_{n=1}^N M(\bfs_n), 
    \label{eq:5}
    \end{equation}
    where $M(\bfs_n)= \frac{w_1 \times (f_{NN_1}(\mathbf{X}_{\phi}(\bfs) | \boldsymbol{\theta})-Z_1(\bfs_n))^2 + w_2 \times (f_{NN_2}(\mathbf{X}_{\phi}(\bfs) | \boldsymbol{\theta})-Z_2(\bfs_n))^2}{2}.$
    and
    $w_u \propto \sigma^{2}_u, u = 1,2 $. We have chosen $w_u = \frac{1}{\sigma^{2}_u}$. Here $\sigma^{2}_u$ is unknown and can be estimated through the sample variance of the u-th variable. The final predictor of the neural network would be the bivariate prediction vector $\mathbf{f}_{NN}^{opt}(\mathbf{X}_{\phi}(\bfs_0))$ where 
    \\
    $$
    \hat{\boldsymbol{\theta}} = \operatorname*{argmin}_{\boldsymbol{\theta}} R\{\mathbf{f}_{NN}(\mathbf{X}_{\phi}(\bfs))| \boldsymbol{\theta},\mathbf{Z}_N\}.
    $$
    
\subsection{The link between Bivariate DeepKriging and LMC}

For the univariate case, \cite{chen2020deepkriging} established a connection between DeepKriging and Kriging. They demonstrated that fixed rank kriging (FRK) \citep{cressie2008fixed} is a linear function of covariates and basis functions, representing a special case of DeepKriging with one layer when all activation functions are set to be linear.

\textbf{Theorem:} For a co-located bivariate spatial process, assuming that the latent variables are constructed with the same sets of basis functions, the Linear Model of Co-regionalization (LMC) represents a special case of BDK.

\textbf{Proof:}
For the bivariate case, the construction of LMC \citep{genton2015cross} provides insights into such a link. If $\mathbf{Y}(\bfs)$ is a zero-mean process then it can simply be represented as the underlying bivariate spatial process $\boldsymbol{\gamma}(\bfs)$ which again can be written as 
\[
\boldsymbol{\gamma}(\bfs) = \mathbf{AU}(\bfs),
\]
where $\mathbf{A} = \begin{bmatrix} a_{11} & a_{12} \\ a_{21} & a_{22} \end{bmatrix}$ and $\mathbf{U}(\bfs) = \{ U_1(\bfs), U_2(\bfs)\}$ with $cov(U_1(\bfs), U_2(\bfs)) = 0$. 

Now, using the Karhunen-Lo\`eve Theorem \citep{adler2010geometry}, we can approximate them as 
\[ U_1(\bfs) = \sum_{b=1}^{K} w_{1,b}\phi_{1,b}(\bfs) \]
and 
\[ U_2(\bfs) = \sum_{b=1}^{K} w_{2,b}\phi_{2,b}(\bfs). \]

We then write
\begin{align*}
    \boldsymbol{\gamma}(\bfs) &= \mathbf{AU}(\bfs) \\
    &= \begin{bmatrix} a_{11} & a_{12} \\ a_{21} & a_{22} \end{bmatrix} \begin{bmatrix} U_1(\bfs) \\ U_2(\bfs) \end{bmatrix} \\
    &= \begin{bmatrix} a_{11}\sum_{b=1}^{K} w_{1,b}\phi_{1,b}(\bfs) + a_{12}\sum_{b=1}^{K} w_{2,b}\phi_{2,b}(\bfs) \\ a_{21}\sum_{b=1}^{K} w_{1,b}\phi_{1,b}(\bfs) + a_{22}\sum_{b=1}^{K} w_{2,b}\phi_{2,b}(\bfs) \end{bmatrix} \\
    &= \begin{bmatrix} \sum_{b=1}^{K} (a_{11}w_{1,b}\phi_{1,b}(\bfs) + a_{12}w_{2,b}\phi_{2,b}(\bfs)) \\ \sum_{b=1}^{K} (a_{21}w_{1,b}\phi_{1,b}(\bfs) + a_{22}w_{2,b}\phi_{2,b}(\bfs)) \end{bmatrix}.
\end{align*}

Now, as both the random variables $U_1(\bfs)$ and $U_2(\bfs)$ are co-located, we can choose the same set of basis functions for them, resulting in 
\begin{align*}
    &= \begin{bmatrix} \sum_{b=1}^{K} (a_{11}w_{1,b} + a_{12}w_{2,b})\phi_{b}(\bfs) \\ \sum_{b=1}^{K} (a_{21}w_{1,b}+ a_{22}w_{2,b}) \phi_{b}(\bfs)\end{bmatrix} , \text{ writing }\phi_{1,b}(\bfs) = \phi_{2,b}(\bfs) = \phi_{b}(\bfs) \\
    &= \mathbf{W}_1\boldsymbol{\phi}(\bfs).
\end{align*}
which is a multioutput regression with $\boldsymbol{\phi}(\bfs)$ as covariates and $\boldsymbol{\gamma}(\bfs)$ as the response, which is same as \eqref{eq:6} with $L = 1$, $\mathbf{b}_1 = \mathbf{0}$, and a linear activation function.

\subsection{Prediction uncertainty}

The utilization of neural networks for uncertainty quantification has garnered significant attention in recent years. Various prominent techniques have emerged \citep{khosravi2011comprehensive,nourani2021prediction} to enhance the estimation of prediction intervals. For instance, \cite{cannon2002downscaling} and \cite{jeong2005rainfall} employed ensemble DNNs for forming prediction intervals, while \cite{srivastav2007simplified} and \cite{boucher2009tools} introduced bootstrap-based uncertainty quantification for DNNs. Additionally, \cite{kasiviswanathan2016comparison} proposed a Bayesian framework for uncertainty quantification with DNNs.

The univariate DeepKriging model computed prediction intervals using a histogram-based approximation to the predictive distribution. Nonetheless, this method exhibits several limitations. Firstly, the accuracy of the prediction intervals heavily relies on the choice of bins for the histogram. The selection of cut points in the histogram, which are randomly chosen from a uniform distribution, may impact the approach's performance. Moreover, the computational time increases when training a large number of ensemble models, and the ensemble approach suggested by this model might fail to capture the variation in the data. In this section, we propose an uncertainty quantification method to address these limitations.

For the bivariate spatial prediction problem, utilizing \eqref{eq:obs} and \eqref{eq:6}, the prediction at an unobserved location $\bfs_0$ can be expressed as
$$
    \hat{\mathbf{Z}}(\bfs_0) =  \mathbf{f}_{NN}^{opt}(\mathbf{X}_{\phi}(\bfs_0)) + {\boldsymbol{\epsilon}}(\bfs_0).
$$

By employing ensembles, we can generate $B$ replications of $\hat{\mathbf{Z}}(\bfs_0)$ at $\bfs_0$. Consequently, the prediction can be articulated as
\begin{equation}
\begin{aligned}
    \hat{\mathbf{Z}}(\bfs_0)^B =& \left(\frac{1}{B} \sum_{i = 1}^B \hat{Z}_1(\bfs_0)_i,\frac{1}{B} \sum_{i = 1}^B \hat{Z}_2(\bfs_0)_i\right)^T \\
    =& \left(\frac{1}{B} \sum_{i = 1}^B \left(f_{NN_1}^{opt}(\mathbf{X}_{\phi}(\bfs_0)) + {{\epsilon}}_1(\bfs_0)\right)_i,\frac{1}{B} \sum_{i = 1}^B \left(f_{NN_2}^{opt}(\mathbf{X}_{\phi}(\bfs_0)) + {{\epsilon}}_2(\bfs_0)\right)_i\right)^T.
\end{aligned}
\label{pred_mean}
\end{equation}
% Note, that here the $B$ ensample models are modeled based on $B$ bootstrap samples. Hence, we can employ the idea of bootstrap-t intervals \citep{hesterberg2011bootstrap} to obtain the prediction bounds. The bootstrap-t interval is thus given as 
% \[\hat{{Z}}_u(\bfs_0)^B \pm t_{(1-\alpha/2),\text{df}} S_{\hat{{Z}}_u(\bfs_0)^B} , \ u = 1,2.\]
% where, $S_{\hat{{Z}}_u(\bfs_0)^B}$ is the standard error which can be further decomposed as
% $S_{\hat{{Z}}_u(\bfs_0)^B} = \text{Var}(Y_u(\bfs_0)) + \text{Var}(\epsilon_u(\bfs_0))$. Assuming independence between $Y_u(\bfs_0)$ and $\epsilon_u(\bfs_0)$, from the outcomes of $B$ ensembles, we can estimate $\text{Var}(Y_u(\bfs_0))$ with $\widehat{{Var}(Y_u(\bfs_0))}$ as
% \begin{equation}
% \widehat{{Var}(Y_u(\bfs_0))} = \frac{1}{B-1} \sum_{i=1}^B f_{NN_u}^{{opt}}(\mathbf{X}{\phi}(\bfs_0))^2_i - \left( \frac{1}{B} \sum_{i=1}^B f_{NN_u}^{{opt}}(\mathbf{X}_{\phi}(\bfs_0))_i\right)^2.
% \label{model_variance}
% \end{equation}
Employing the multidimensional Central Limit Theorem \citep{Multi_CLT} yields $\hat{\mathbf{Z}}(\bfs_0)^B \sim \mathcal{N}_2(\mathbf{Z}(\bfs_0), \frac{1}{B}\Sigma_{\mathbf{Z}(\bfs_0)})$. The prediction interval is thus given as 
\[\hat{{Z}}_u(\bfs_0)^B \pm z_{(1-\alpha/2)} \sigma(Z_u(\bfs_0)) , \ u = 1,2,\]
where, $\sigma^2(Z_u(\bfs_0))$ is the variance term associated with $Z_u(\bfs_0), \ \text{for } u = 1,2$, which can be decomposed as $\sigma^2(Z_u(\bfs_0)) = \text{Var}(Y_u(\bfs_0)) + \text{Var}(\epsilon_u(\bfs_0))$, assuming independence between $Y_u(\bfs_0)$ and $\epsilon_u(\bfs_0)$. From the outcomes of $B$ ensembles, we can express $\widehat{{Var}(Y_u(\bfs_0))}$ as
\begin{equation}
\widehat{{Var}(Y_u(\bfs_0))} = \frac{1}{B-1} \sum_{i=1}^B f_{NN_u}^{{opt}}(\mathbf{X}{\phi}(\bfs_0))^2_i - \left( \frac{1}{B} \sum_{i=1}^B f_{NN_u}^{{opt}}(\mathbf{X}_{\phi}(\bfs_0))_i\right)^2.
\label{model_variance}
\end{equation}

The noise variance $Var(\epsilon_u(\bfs_0))$ can then be represented as
$$
\begin{aligned}
    Var(\epsilon_u(\bfs_0)) =& \text{Total Variance} - \text{Model variance} \\
    =& \mathbb{E}\{(Z_u(\bfs_0) - \hat{Z}_u(\bfs_0))^2\} - \widehat{Var(Y_u(\bfs_0))}.
\end{aligned}
$$
An empirical approximation of the above is
\begin{equation}
r_u^2(\bfs_0) = \max\{(Z_u(\bfs_0) - \frac{1}{B} \sum_{i=1}^B f_{NN_u}^{\text{opt}}(\mathbf{X}_{\phi}(\bfs_0))_i)^2 - \widehat{\text{Var}(Y_u(\bfs_0))}, 0\}, \ u = 1,2.
\label{r}
\end{equation}
Consequently, based on the above formulation, the prediction interval at the $\alpha$ level of significance for variable $u$ at location $\bfs_0$ can be given as,
\begin{equation}
\hat{{Z}}_u(\bfs_0)^B \pm z_{(1-\alpha/2)} \sqrt{\frac{1}{B} \left(\widehat{\text{Var}(Y_u(\bfs_0))} + r_u^2(\bfs_0) \right)} , \ u = 1,2
\label{pred_interval}
\end{equation}
where $z_{(1-\alpha/2),\text{df}}$ represents the $1-\alpha/2$ quantile of the Normal distribution.

In connection with the preceding explanation, we now provide a comprehensive definition of the algorithm for calculating the prediction interval. We partition our training dataset $D$ into two subsets, $D_1$ and $D_2$. From $D_1$, we extract a sample $D_{11}$. Initially, we train a neural network model with $L$ layers using the dataset $D_{11}$, and then proceed to draw $B$ bootstrap samples from $D_1$ to train $B$ DNN models. For these $B$ bootstrap fits, we keep the weights fixed for the first $L_0$ layers, allowing only the new models to be trained based on the last $L-L_0$ layers. This approach is adopted to control the degrees of freedom of the t-distribution in \eqref{pred_interval}. By exclusively re-training the last $L-L_0$ layers, we ensure that the total number of trainable parameters $p$ in each bootstrap computation remains smaller than $N$. Following this procedure, we compute $\hat{Z}_u(\bfs_0)$ and $\widehat{Var}(Y_u(\bfs_0))$ based on the training data $D_1$. Since we know $\mathbb{E}{\epsilon_u(\bfs_0)} = 0$, we have $\hat{Z}u(\bfs_0) = \frac{1}{B} \sum_{i = 1}^B f_{NN_u}^{{opt}}(\mathbf{X}_{\phi}(\bfs_0))$.

Computation of $r_u^2(\bfs_0)$ as defined in \eqref{r} is infeasible as we do not have $Z_u(\bfs_0)$. Hence we estimate $r_u^2(\bfs_0)$ through the second part of the training dataset $D_2$. To do this we use the nearest neighbour approach. 
We obtain a set $D_{20} = \{\bfs_k: \bfs_k \in D_{2}\}$ with cardinality $G$ of locations closest to $\bfs_0$ and take average of the corresponding $r_u^2(\bfs_k)$'s, i,e.,
\begin{equation}
    \hat{r}_{u}^2(\bfs_0) = \frac{1}{G}\sum_{\bfs_g \in D_{20}} r_u^2(\bfs_g), 
    \label{eq:14}
\end{equation}
such that $\bfs_g$'s are the nearest $G$ locations to $\bfs_0$.

% Now at $\alpha$ level of significance we can write the prediction interval at $\bfs_0$ as 
% \begin{equation}
%     \hat{\mathbf{Y}}_{mean_u}(\bfs_0) \pm t_{(1-\alpha/2),df} \sqrt{\hat{\Sigma}_{\bfs_0_{uu}} + \hat{\sigma}_{\epsilon_u}^2(\bfs_0) } , \ u = 1,2
%     \label{eq:9}
% \end{equation}
% where $t_{(1-\alpha/2),df}$ is the $1-\alpha/2$ quantile of $t$-distribution with $df$ degrees of freedom, $df=N-p$, where $p$ is the Number of estimated parameters. Note that for these bootstrap neural networks we have trained only $L - L_0$ layers resulting in reduced number of parameters in the modelling. This ensures $df$ to be positive.

Algorithm \ref{algo:1} provides a step-by-step explanation of our prediction interval construction method as discussed in previous paragraph.

\begin{algorithm}
    \caption{Prediction Intervals Algorithm}\label{alg:cap}
    \begin{algorithmic}
    \State Split $\mathbf{D}$ into $\mathbf{D_1}$ and $\mathbf{D_2}$ equally.
    \State Further split $\mathbf{D_1}$ into $\mathbf{D_{11}}$ and $\mathbf{D_{12}}$.
    \State Train a deep neural network (\textbf{DNN}) of $L$ layers on $\mathbf{D_{11}}$.
    \State Take \textbf{B} random samples $\{\mathbf{D_{1}^{1},D_{1}^{2},...,D_{1}^{B}}\}$ from $\mathbf{D_1}$.
    
    \For{$i \gets 1$ to \textbf{B}}   
            \State Fix the weights of the first $L_0$ layers of the \textbf{DNN} and train the last $L-L_0$ layers on $\mathbf{D_{1}^{i}}$.
            \State Train the \textbf{DNN} on $\mathbf{D_{1}^{i}}$ and store the result. Denote it as $\mathbf{f}_{NN}^{opt}(\mathbf{X}_{\phi}(\bfs))_i$.
    \EndFor
    
    \For{location $\bfs_k$ in $\mathbf{D_2}$} 
            \State Calculate $\hat{{Z}}_u(\bfs_k)^B$\eqref{pred_mean} and $\widehat{{Var}(Y_u(\bfs_k))}$ \eqref{model_variance}.
            \State Calculate ${r}_{u}^2(\bfs_k)$ \eqref{r}.
    \EndFor
    
    \State For test location $\bfs_0$, obtain the set $\mathbf{D_{20}} = \{\bfs_k: \bfs_k \in \mathbf{D_{2}}\}$ of the nearest $G$ locations from $\bfs_0$.
    
    \State Calculate $\hat{{Z}}_u(\bfs_0)^B$\eqref{pred_mean}, $\widehat{{Var}(Y_u(\bfs_k))}$ \eqref{model_variance}, and $\hat{r}_{u}^2(\bfs_0)$ \eqref{eq:14}.
    
    \State Calculate the prediction interval as defined in \eqref{pred_interval}.\\
    \Comment{Where $u$ stands for the $u$-th variable, $u = 1,2$.}
    \end{algorithmic}
    \label{algo:1}
\end{algorithm}

\subsection{Computational scalability}

To better understand the computational benefits of DNN over traditional kriging we can look at the time complexity of the DNN with kriging. DeepKriging involves matrix multiplication in several layers to give us the resulting output. Whereas, kriging involves a $2N \times 2N$ matrix inversion. Note that, time complexity of multiplication of one $m \times n$ and $n \times p$ matrix is $O(mnp)$. So a single layer $l$ of the neural network with minibatch (a minibatch \citep{hinton2012neural} is a randomly selected portion of the data used for neural network training.) size of $b$, $M_{l-1}$ input nodes and $M_{l}$ output nodes will have the time complexity of $O(bM_{l-1}M_{l})$.  Faster training is made possible by defining a minibatch. Hence a neural network with $L$ layers will have time complexity $O(\sum_{l=1}^{L}bM_{l-1}M_{l})$. On the other hand, time complexity of Kriging is $O(8N^3)$. Hence for large $N$ deep kriging with adequate number of layers and nodes is more computationally efficient than traditional kriging.

\section{Simulation studies}\label{sec:simulation}

\subsection{Point predictions}\label{subsec:point_prediction}
We have conducted several experiments through Gaussian, non-Gaussian and nonstationary simulations to give a comprehensive comparison of our proposed method with the parsimonious bivariate Mat\'ern (CMK) as well as the linear model of coregionalization (CLK). In order to compare our results we have computed the Root Mean Square Prediction Error for each variable as the validation metric. For $N$ locations this can be  given as 
$$
    {MSPE}_u =  {\frac{1}{N}\sum_{n=1}^N (Z_u(\mathbf{s_n}) - \hat{Z_u}(\mathbf{s_n}))^2}, \ u = 1,2. 
$$

Our model performed at par with CMK and CLK in the Gaussian scenario. Details of the study has been provided in the supplimentary materials. In this section we mainly concentrate on non-Gaussian and nonstationary scenarios where BDK clearly outperforms the traditional statistical techniques. 

Here we consider two simulation settings. The first one is  bivariate non-Gaussian with covariates. We simulated the model from the following framework
$$
\mathbf{Y}(\bfs) = \boldsymbol{\mu}(\bfs) + \boldsymbol{\gamma}(\bfs), 
$$
where $\boldsymbol{\mu}(\bfs)$ is the mean function and $\boldsymbol{\gamma}(\bfs)$ follows a non-Gaussian distribution. We define $\boldsymbol{\mu}(\bfs) = \{\mu_1(\bfs),\mu_2(\bfs)\}$ as a complex function of covariates given as:
$$
\begin{aligned}
    {\mu_u}(\bfs) =& x_1(\bfs)^2 - x_2(\bfs)^2 + x_3(\bfs)^2 - x_4(\bfs)^2 - \\
&x_5(\bfs)^2 + 2x_1(\bfs)x_2(\bfs) + 3x_2(\bfs)x_3(\bfs) - 2x_3(\bfs)x_5(\bfs) + \\
&10x_1(\bfs)x_4(\bfs) + \sin(x_1(\bfs))x_2(\bfs)x_3(\bfs) + \\
&\cos(x_2(\bfs))x_3(\bfs)x_5(\bfs) + (x_1(\bfs)x_2(\bfs)x_4(\bfs)x_5(\bfs)), u = 1,2
\end{aligned}
$$
where the covariates $x_i(\bfs),i = 1,\ldots,5$ were generated from multivariate Gaussian distributions with Mat\'ern covariance with varying parameters. $\boldsymbol{\gamma}(\bfs)$ was simulated from a multivariate Gaussian distribution with parsimonious Mat\'ern covariance \eqref{eq:2} with parameters $\sigma_1 = 0.7, \sigma_2 = 0.8, \nu_1 = 0.3, \nu_2 = 0.6, \alpha_1^2 = 0.05, \alpha_2^2 = 0.1, R = 0.5$. We generated $100$ replicates of the random field at 1200 irregular locations on $[0,1] \times [0,1]$ by transforming the Gaussian fields from the previous simulation using the Tukey-g and h transformation \citep{xu2017tukey} defined as 
$$\tau_{g,h}(z) = g^{-1}\{exp(g z) - 1\} exp(h z^2/2),$$ with $g = 0.8,h = 0.5$
for variable 1 and $g=-0.8,h = 0.5$ for variable 2.  

\begin{figure}[ht]
    \centering
   \includegraphics[scale=0.5]{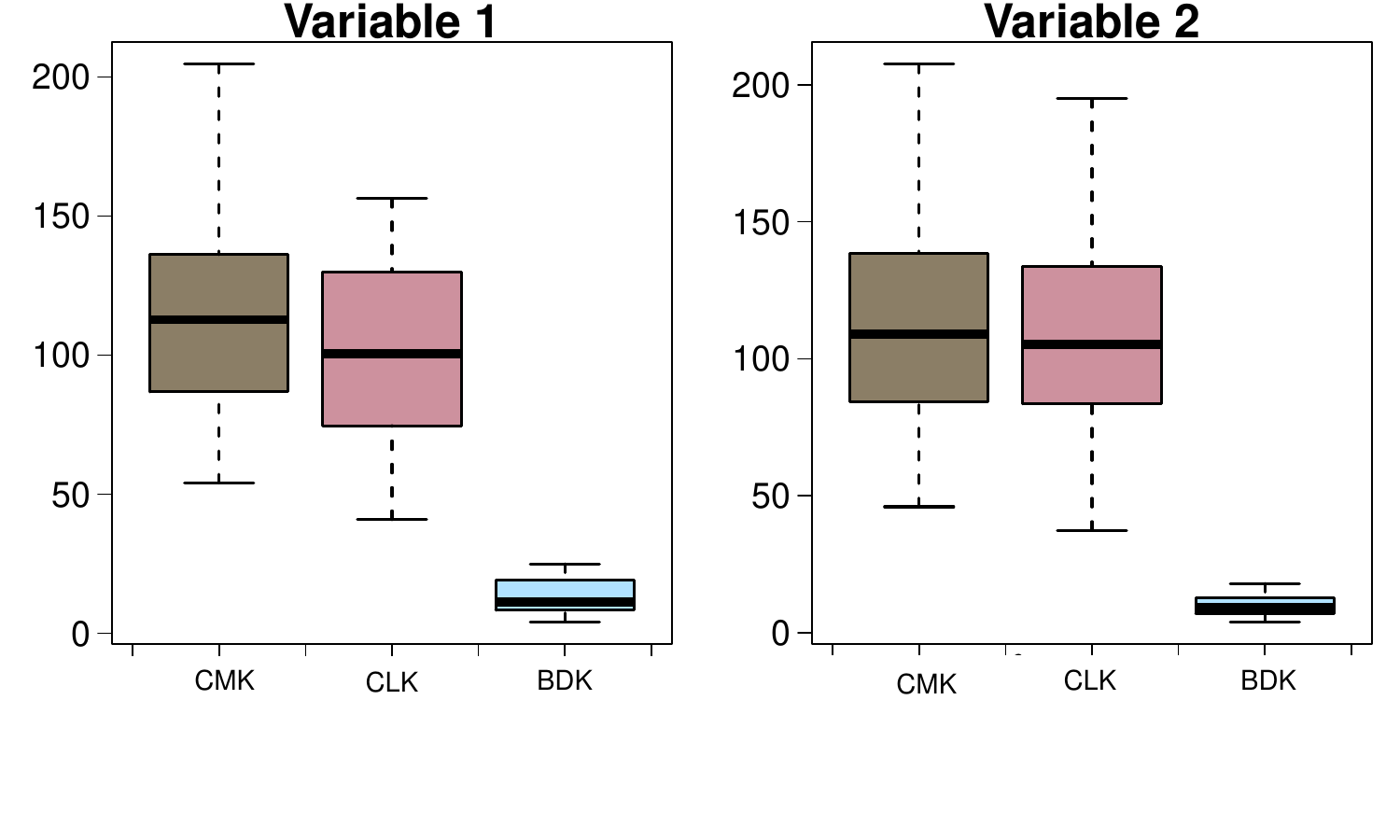}
   \caption{Boxplot of MSPE for the three different models CMK, CLK and BDK over $100$ replicates for the non-Gaussian simulation.}
    \label{fig:2}
\end{figure}

Figure \ref{fig:2} illustrates the boxplot of MSPE for the two variables across $100$ replicates. The superiority of BDK over other conventional approaches is evident from the boxplots. It's worth noting that due to parameter configurations and data sample paths, optimization in CMK and CLK may present challenges. In contrast, BDK is devoid of such limitations and can be readily implemented on any dataset.

The second simulation explores nonstationary datasets. We simulate nonstationary random fields using the Wendland radial basis functions as discussed in Section \ref{sec:deepkriging}. We apply a nonlinear transformation to the basis functions to obtain the nonstationary fields, defined as 
\begin{align*}
Y_1(\mathbf{s}) &= \sum_{i=1}^{19}\left((a \cdot \phi_{2i}(\mathbf{s})^{3/2} + c \cdot \phi_{2i-1}(\mathbf{s})) - b \cdot \sqrt{\phi_{2i}(\mathbf{s}) \cdot \phi_{2i-1}(\mathbf{s})}\right) \\
Y_2(\mathbf{s}) &= \sum_{i=1}^{19}\left((d \cdot \phi_{2i}(\mathbf{s}) - e \cdot \phi_{2i-1}(\mathbf{s})^{3/2})\right)
\end{align*}
where parameters $\{a,b,c,d,e\} \sim \text{Uniform}(-2.5,2.5)$.

\begin{figure}[!htb]
    \centering
   \includegraphics[scale=0.5]{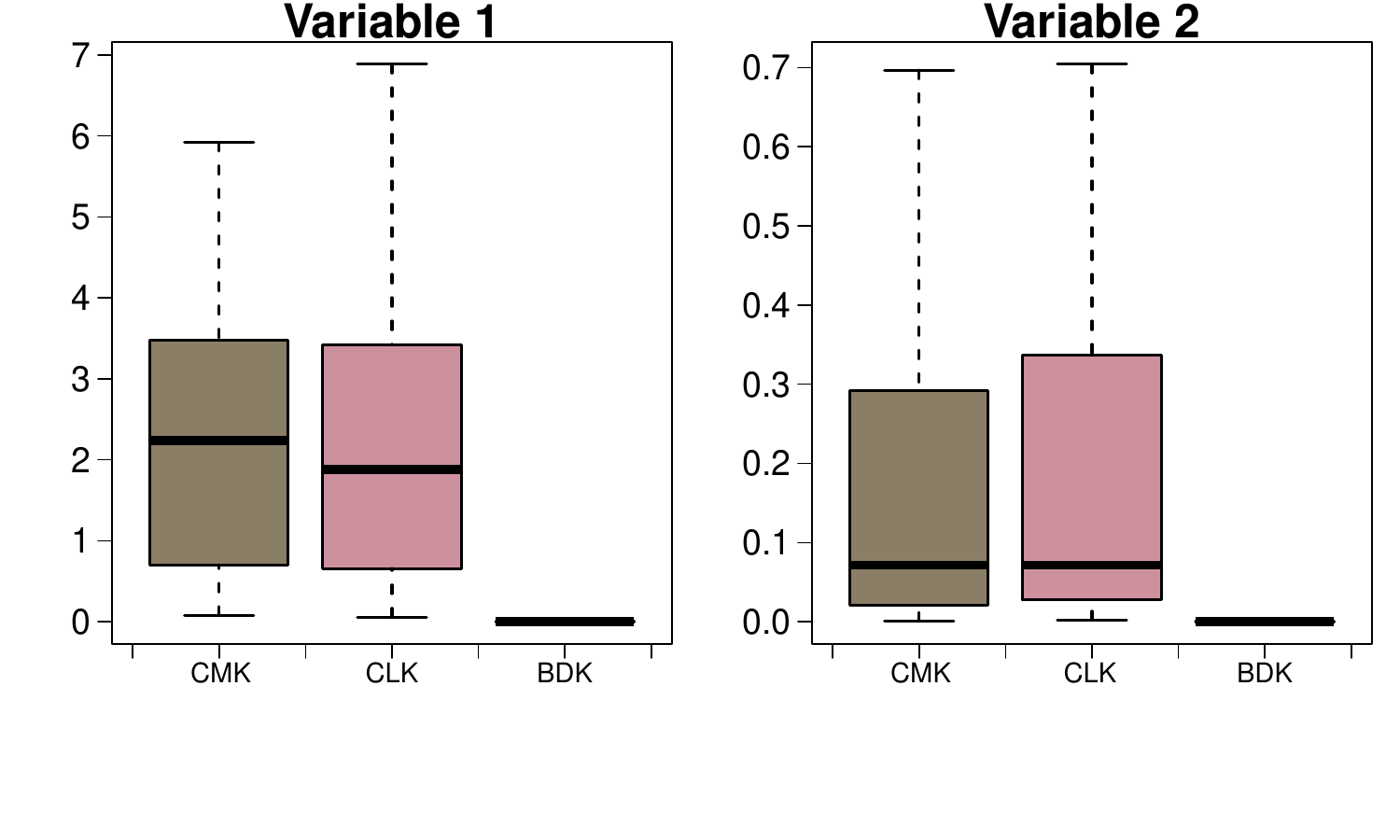}
   \caption{Boxplot of MSPE for the three different models CMK, CLK and BDK over $100$ replicates for the nonstationary simulation.}
    \label{fig:nonstat_boxplot}
\end{figure}

Figure \ref{fig:nonstat_boxplot} displays the RMSPE over $100$ replicates. It is evident that BDK significantly outperforms the other comparative methods.

% A visual plot of the data is given in figure \ref{fig:2}.

% \begin{figure}[h]
%     \centering
%   %\includegraphics[scale=0.28]{ImportanceBA.pdf}
%   \includegraphics[scale=0.4]{images/nongaussian_1200.pdf}
%   \caption{Visualization of the simulated nongaussian data as an image.}
%     \label{fig:2}
% \end{figure}

\subsection{Prediction intervals}

Several studies \citep{ho2001neural,zhao2008statistical} have proposed various approaches for quantifying the quality of the prediction interval. We have chosen here two specific measures for comparison of the prediction bands, namely prediction interval coverage probability (PICP) and mean prediction interval width (MPIW). In mathematical notations
    \begin{equation}
    \begin{aligned}
        PICP_u = \frac{1}{N}\sum_{j=1}^N \mathbf{1}_{Z_u(s_j)\in [L_u(s_j),U_u(s_j)]}, \ 
        MPIW_u = \frac{1}{N}\sum_{j=1}^N [U_u(s_j) - L_u(s_j)], \ u=1,2
    \end{aligned}
    \label{eq:16}
    \end{equation}
    where $(L_u(s),U_u(s))$ are the lower and upper prediction bounds for variable $u$ at location $\bfs$ and $n$ is the number test samples.

In this section we evaluate our proposed approach of the prediction interval computation with the parametric prediction intervals obtained from cokriging models. We have computed the prediction interval validation metrics \eqref{eq:16} for the comparative models using the same set of simulations as defined in \ref{subsec:point_prediction}. In all of the following accessments we have computed the $95\%$ prediction bound for predictions in the test set. 

Table \ref{tab:1} shows the results for all the simulation scenarios. It can be seen that for the Gaussian simulations where ${CMK}_{true}$ is optimal the BDK model gives comparable performance. For non-Gaussian and non-stationary scenarios clearly the BDK outperforms the other methods by attaining the $95\%$ prediction with smaller width. Note that, some of the simulation scenarios 

\begin{table}[!htb]
    \centering
    \caption{Comparison on both the variables over different simulation settings.}
    \resizebox{\columnwidth}{!}{%
    \begin{tabular}{||c c c c c c c c c c||} 
        \hline
        Simulation type & Models & $MSPE_1$ &${SE}_1$ & ${PICP}_1$  & ${MPIW}_1$  & $MSPE_2$ & ${SE}_2$ & ${PICP}_2$  & ${MPIW}_2$ \\ [0.5ex] 
         \hline\hline
         Gaussian & 
         \begin{tabular}{c} $CMK_{true}$\\
         BDK\\ \end{tabular} & 
         \begin{tabular}{c}
         {0.23}\\ 
         0.24\\
         \end{tabular} &
         \begin{tabular}{c}
         {0.11}\\ 
         0.19\\
         \end{tabular} &
         \begin{tabular}{c}
         0.95\\ 
         0.94 \\
         \end{tabular} &
         \begin{tabular}{c}
         0.87\\ 
         0.98 \\
         \end{tabular} &
         \begin{tabular}{c} 
         {0.21}\\ 
         0.24 \\
         \end{tabular} &
         \begin{tabular}{c}
         {0.09}\\ 
         0.18\\
         \end{tabular} &
         \begin{tabular}{c} 
         0.95 \\ 
         0.95 \\
         \end{tabular} &
         \begin{tabular}{c} 
         0.92 \\ 
         1.07 \\
         \end{tabular} \\
         \hline
         non-Gaussian & 
         \begin{tabular}{c} CMK\\
         CLK \\
         BDK\\ \end{tabular} & 
         \begin{tabular}{c}
         12.1 ($\times 10$)\\
         87.4\\
         32.7\\
         \end{tabular} &
         \begin{tabular}{c}
         0.29 ($\times 10$)\\
         12.13\\
         11.6\\
         \end{tabular} &
         \begin{tabular}{c}
         0.27\\ 
         0.58\\
         0.94 \\
         \end{tabular} &
         \begin{tabular}{c}
         6.21\\
         11.2\\
         29.9 \\
         \end{tabular} &
         \begin{tabular}{c} 
         12.4 ($\times 10$)\\
         94.5\\
         23.8 \\
         \end{tabular} &
         \begin{tabular}{c}
         0.49 ($\times 10$)\\
         21.9\\
         9.11\\
         \end{tabular} &
         \begin{tabular}{c} 
         0.26 \\ 
         0.51 \\ 
         0.94 \\
         \end{tabular} &
         \begin{tabular}{c} 
         6.16 \\
         9.99 \\
         29.4 \\
         \end{tabular} \\
         \hline
         non-stationary & 
         \begin{tabular}{c} CMK\\
         CLK \\
         BDK\\ \end{tabular} & 
         \begin{tabular}{c}
         1.95\\ 
         1.26\\ 
         7.52 ($\times 10^{-4}$)\\
         \end{tabular} &
         \begin{tabular}{c}
         0.75\\ 
         0.14\\ 
         1.01 ($\times 10^{-4}$)\\
         \end{tabular} &
         \begin{tabular}{c}
         0.92\\ 
         0.92\\ 
         0.96 \\
         \end{tabular} &
         \begin{tabular}{c}
         3.53\\ 
         3.49\\
         0.16 \\
         \end{tabular} &
         \begin{tabular}{c} 
         {0.13}\\ 
         0.14\\ 
         6.83 ($\times 10^{-4}$) \\
         \end{tabular} &
         \begin{tabular}{c}
         0.02\\ 
         0.02\\ 
         1.08 ($\times 10^{-4}$)\\
         \end{tabular} &
         \begin{tabular}{c} 
         0.09 \\ 
         0.10 \\
         0.95 \\
         \end{tabular} &
         \begin{tabular}{c} 
         1.01 \\
         1.33 \\
         0.19 \\
         \end{tabular} \\
         \hline
    \end{tabular}%
    }
    \label{tab:1}
\end{table}

\begin{figure}[!htb]
    \centering
   \includegraphics[scale=0.63]{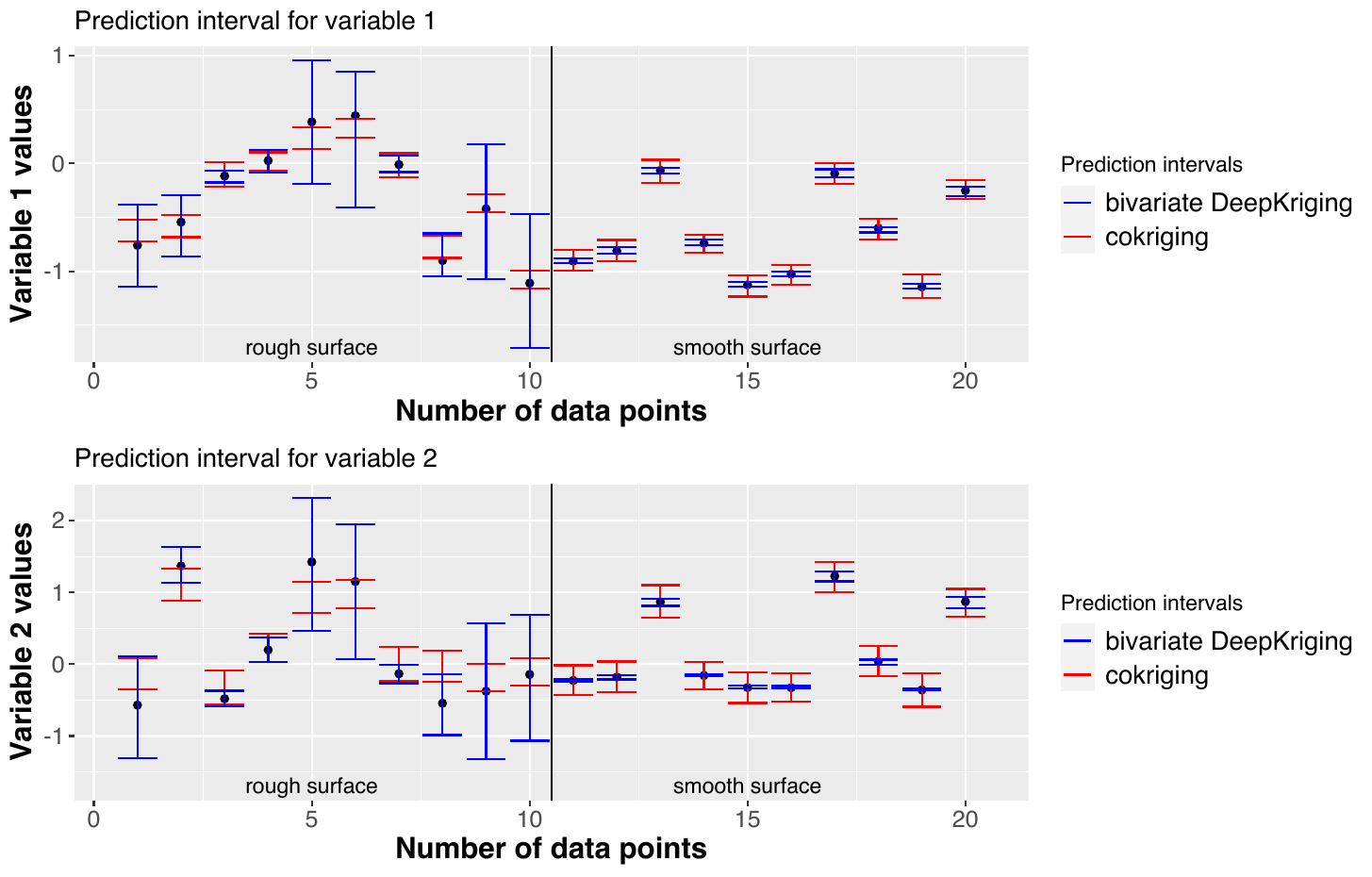}
   \caption{Prediction interval for variable 1 and variable 2 for the nonstationary simulation. Here the red interval signifies the prediction interval generated by CMK and the blue interval is for BDK. }
    \label{fig:pred_interval_nonstat}
\end{figure}

While Section \ref{subsec:point_prediction} demonstrates how the nonlinearity in BDK aids in capturing the nonstationarity in the data, it's equally important to showcase the improvement in prediction intervals offered by the proposed BDK model. To investigate this, we generate nonstationary data with a nonstationary mean function. The simulation details are provided in supplementary materials. 

Figure \ref{fig:pred_interval_nonstat} displays the prediction intervals for BDK and the cokriging models for the nonstationary simulation. The prediction intervals for BDK adapt spatially. Specifically, BDK offers wider prediction bounds for regions with higher variance, whereas it provides narrower prediction bounds for regions with lower variability.

\subsection{Computation time}\label{sec:Comp_time}

Based on the same simulation setting we investigate the computational time of BDK compared to CMK with different number of observations N. We have fully optimized the model and have measured the computation time for the whole process in each case. We have also compared CMK with both of our proposed methods for point prediction as well as interval prediction.
\begin{figure}[!htb]
    \centering
   \includegraphics[width=\textwidth]{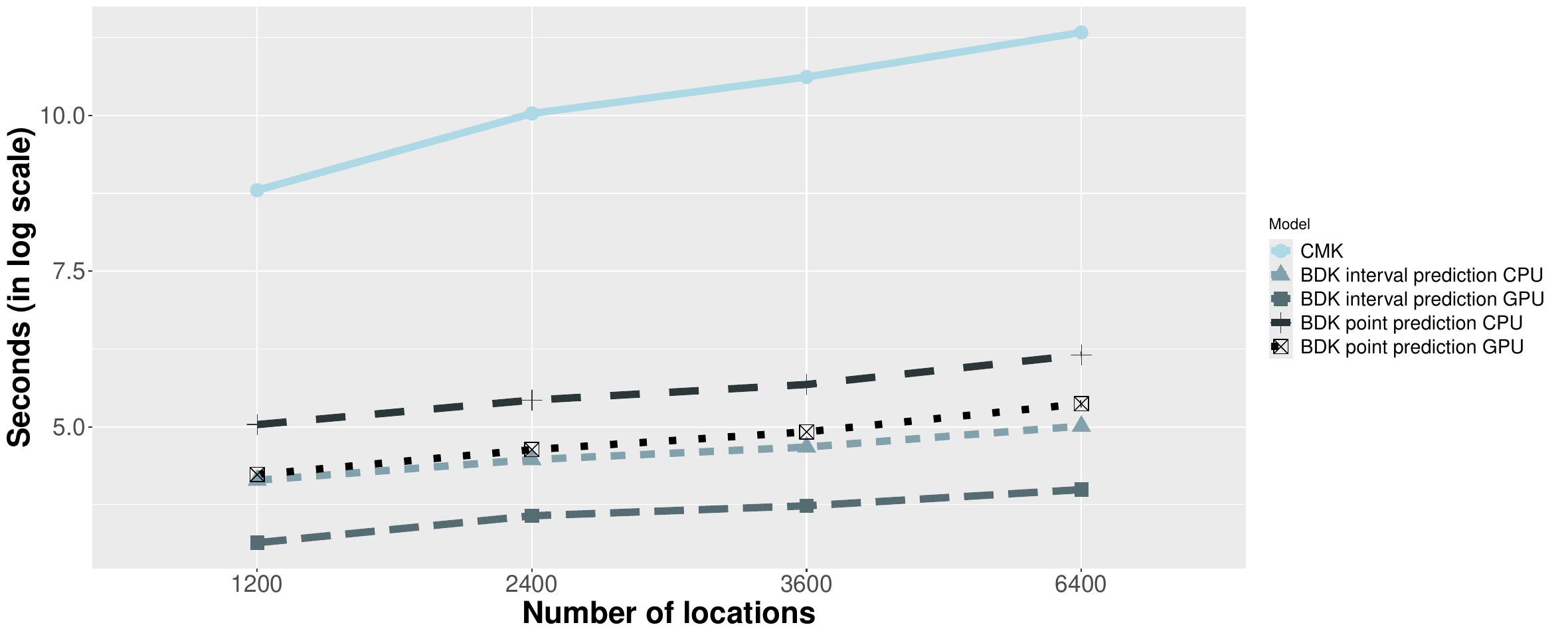}
   \caption{Total computation time (in secconds) for different models in log scale for different number of locations}
    \label{fig:11}
\end{figure}
Figure \ref{fig:11} shows the comparison results. As it can be seen BDK always outperforms CMK in computational time hence it is much more scalable when the sample size increases. For example, when N = 6400, which is the largest sample size we have considered, it takes more than 23 hours (83,484 seconds) to train the cokriging model, which makes it computational infeasible for larger N. However, for the same data size, BDK with prediction interval only costs 7.85 minutes (471 seconds) without GPU acceleration and 3.58 minutes (215 seconds) with a Tesla P100 GPU. We used the GPU from Ibex a shared computing platform at KAUST. For a more powerful GPU, the computational cost will be further reduced.

\section{Application}\label{sec:application}

In recent years, Saudi Arabia is seeking to reduce its reliance on fossil fuels for energy demand by investing in renewable energy sources. To achieve this goal, they have introduced several milestones for their upcoming smart cities. For King Abdullah City for Atomic and Renewable Energy (KA-CARE, 2012) they are planning for 54 GW of renewable energy portfolio by 2032, of which 9 GW will come from wind power. The Saudi Vision 2030 (2016) aims at achieving 9.5 GW of renewable energy by 2023. NEOM, an acronym for "New Future" and "New Enterprise Operating Model", is the upcoming mega-city project for Saudi Arabia which is planned to use only renewable power sources (wind and solar). For instance, wind turbines are widely used throughout the world to produce wind energy. However, installing such turbines necessitates precise measurements of the local wind direction and speed. To achieve this goal, an accurate wind speed interpolation model is crucial.

The majority of previous publications modeled wind data using geostatistical methods.
Kriging was employed by \cite{wang2020wind} to model wind fields. Large wind datasets were modeled for interpolation by \cite{9397281} using the ExaGeoStat software, which uses bivariate parsimonious Mat\'ern covariance structure. However, these techniques always presumptively assume that the data are stationary and Gaussian. In this paper we propose a novel methodology to model the bivariate wind speed which can then be used for spatial prediction and get high resolution maps of wind fields in any particular region. For our application we have also provided a high resolution wind interpolation for the NEOM region. 

We have fitted BDK on the wind data discussed in Supplementary materials. We have used 3 layers of basis functions with $K=100,361,1369$ respectively. The neural network consists of $L=7$ layers with $M_1 = M_2 = M_3 = M_4 = 100$ for first
4 layers, $M_5 = 50, M_6 = 50$ and $M_7=2$ nodes in the final layer. By using a sample from a uniform distribution, the model weights were initialized, and a learning rate of 0.01 was selected. The dataset consists of 506,771 locations. However, due to computational issues, the cokriging models could only handle a small subsample of the data. For this reason, we have randomly chosen $147,000$ locations for training of the cokriging model. On the other hand, the BDK can easily handle the full data. Hence we have fitted two different sets of models with deep neural networks, first with $147,000$ ($BDK_{147000}$) and second with $450,000$ ($BDK_{450000}$) locations and we randomly chosen $1000$ locations from the rest for testing. To make the cokriging model more competitive and to reduce the computational burden, we have splitted the data into 100 sub-regions and fitted the model on each sub-region independently.

Table \ref{tab:real_data} shows that BDK outperforms CMK in MSPE.
\begin{table}[h]
    \centering
    \caption{{MSPE} for CMK and BDK.}
    \begin{tabular}{||c c c||} 
    \hline
    Models & ${MSPE}_1$ & ${MSPE}_2$ \\ [0.5ex] 
    \hline\hline
    CMK & 0.882 & 4.066 \\
    \hline
     $BDK_{147000}$ & 0.488 & 0.438  \\ 
    \hline
    $BDK_{450000}$ & \textbf{0.394} & \textbf{0.392}  \\ 
    \hline
    \end{tabular}
    
    \label{tab:real_data}
\end{table}

Next we look at the 95\% prediction intervals provided by BDK and CMK. From Table \ref{tab:10} we can see that even though the MPIW for CMK is smaller than BDK, it fails drastically containing the true values within its bounds. 

\begin{table}[h]
    \centering
    \caption{PICP and MPIW with CMK and  BDK for variable 1 and variable 2 respectively.}
        \begin{tabular}{||c c c c c||} 
        \hline
        Models & $PICP_1$ & $PICP_2$ & $MPIW_1$ & $MPIW_2$\\ [0.5ex] 
         \hline\hline
          CMK & 0.601 & 0.734 & 1.671 & 1.343 \\
         \hline
         BDK & \textbf{0.971} & \textbf{0.950} & \textbf{1.226} & \textbf{1.340} \\
         \hline
         \end{tabular}
    \label{tab:10}
\end{table}

We have also captured the total computation time using BDK and CMK. For CMK total computation time was 2.18 days (188,352 seconds) where as for BDK it took \textbf{16.81 minutes} (1009 seconds) for point prediction and \textbf{55.61 minutes} (3337 seconds) for interval prediction. 

\subsection{Spatial interpolation of wind field data}

We have given a high resolution interpolation (from $ 5km \times 5km $ to $ 1km \times 1km $) for the region near NEOM, an upcoming smart city in Saudi Arabia. This downscaling can improve our understanding of the wind patterns. \cite{sanchez2020effect} have explored how the change in wind direction affect wind turbine performance. 
\begin{figure}[]
    \centering
   \includegraphics[width=\textwidth]{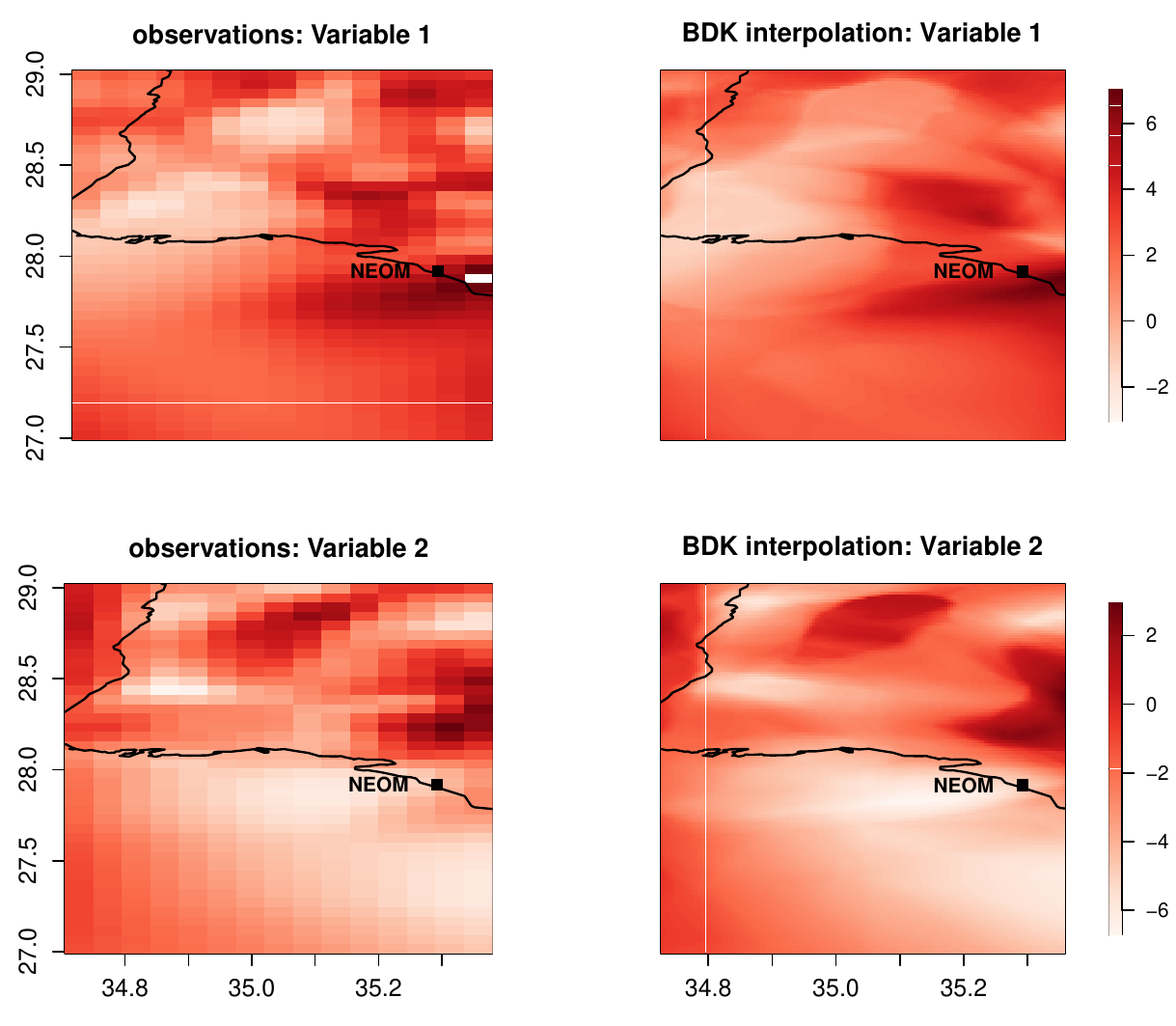}
   \caption{Spatial downscaling from resolution of $5km \times 5km$ to $1km \times 1km$ of the U and V component of wind over the region NEOM }
    \label{fig:10}
\end{figure}
Hence understanding the exact wind patterns in a high resolution field will help in wind energy setup in the area. Figure \ref{fig:10} gives a visual representation of the spatial interpolation carried out using BDK.

\section{Discussion}\label{sec:discussion}

In this work, we have extended the univariate DeepKriging to bivariate scenario for spatial prediction without any parametric assumption of the underlying distribution of the data. We have also introduced a nonparametric approach for prediction interval computation for this deep learning based spatial modelling architecture. Our model is generally compatible with
non-stationarity, non-linear relationships, and non-Gaussian data.

In our implementation we have chosen radial basis functions for spatial embedding. However for long range spatial dependence and circular covariance structures choice of other basis functions such as smoothing spline basis functions \citep{wahba1990spline}, wavelet basis functions \citep{vidakovic2009statistical} can be explored. 

The loss function used to BDK is a weighted MSE where weights are proportional to the variance of the variables. More complex weights can be incorporated to capture complex terrain behaviours. In our study we have not considered topographical elevation. This can be in incorporated by using relevant covariates or by creating an adaptve loss function. 

An interesting avenue of future work for prediction intervals is to extend the conformal prediction methods of \cite{mao2020valid} to the deepKriging case.  This method is model free and thus would avoid the assumption that the prediction intervals are a symmetric t-distribution.

In environmental modeling, interpretability plays a crucial role. When employing deep neural network architectures, one straightforward approach to interpretability involves examining the weight matrix of the input layer. This matrix often provides a basic understanding of the significance of specific covariates. For instance, if the weights corresponding to certain covariates are close to zero, it suggests that their influence on the model output is minimal. Recent developments in interpretability include techniques such as Shapley values \citep{merrick} and Local Interpretable Model-Agnostic Explanations (LIME)  \citep{zafar}. \cite{Zhang2023InterpretableAN} provides an Interpretable architecture for neural networks. We have furnished an extensive review on Shapley values in the supplementary materials, demonstrating how they can aid in understanding the importance of certain covariates.

BDK can be suitable for any large-scale environmental application where high resolution spatial interpolation is a requirement. 

\section*{Supplementary Materials}

\textbf{PDF Supplement:} This document provides comprehensive details about the wind dataset, including initial exploratory analyses. It also presents additional findings on the explainability of neural networks through the application of Shapley values. The PDF file is available as part of the supplementary materials. \\
\textbf{Code Repository:}The supplementary materials contain Python scripts for implementing the proposed method and R scripts for comparative analysis. These resources ensure the reproducibility of all results presented in the article. 
Additionally, the GitHub repository includes the code along with supplementary scripts. You can access it at: \url{https://github.com/pratiknag/Bivariate_DeepKriging/tree/main}

\section*{Acknowledgments}

The authors sincerely thank the Editor, Associate Editor, and two anonymous reviewers for their valuable feedback and constructive suggestions, which have significantly improved the quality of this manuscript.

\section*{Disclosure Statement}

The authors confirm that they have no competing interests to declare.

\bibliographystyle{chicago}

\bibliography{Bibliography-MM-MC}

\end{document}

\if0\blind
{
  \title{\bf Supplementary Material: Bivariate DeepKriging for Large-Scale Spatial Interpolation of Wind Fields}
    \author{Pratik Nag\\
    School of Mathematics and Applied Statistics, \\ University of Wollongong, Australia. \\
    Ying Sun \\
    CEMSE Division, Statistics Program, \\
    King Abdullah University of Science and Technology,Saudi Arabia.\\
    and \\
    Brian J Reich \\
    Department of Statistics, \\
    North Carolina State University, Raleigh, USA.}
  \maketitle
} \fi

\if1\blind
{
  \bigskip
  \bigskip
  \bigskip
  \begin{center}
    {\LARGE\bf Title}
\end{center}
  \medskip
} \fi

\begin{description}

\item[Exploratory analysis of the wind field data:] \label{S1}

Wind field estimates play a fundamental role in many applications, such as power generation \citep{tong2010wind}, hydrological modelling \citep{topalouglu2018analysis,millstein2022can} and monitoring and predicting weather patterns and global climate \citep{arain2007use,bengtsson1988integration}. So it is of great interest to get accurate interpolation of wind fields at unobserved locations to exploit the huge potential of wind data for different fields. Wind is generally quantified as a bivariate variable with two components, $U$ and $V$, where $U$ is the zonal velocity, i.e., the component of the horizontal wind towards east and $V$ is the meridional velocity, i.e., the component of the horizontal wind towards north. The wind speed at any location can then be defined as $\sqrt{U^2 + V^2}$. 

We analyze wind at $5 km \times 5 km$ spatial resolution for January, 2009 (\ref{fig:12}) from a Weather Research and Forecasting (WRF) model simulation on the $[30^{\circ}E,65^{\circ}E] \times [5^{\circ}N,35^{\circ}N]$ region of Earth \citep{yip2018statistical} which spans the Middle East. Figure \ref{fig:12} shows non-stationarity, e.g., the surface is more rough over land than water. The scatter plot in Figure \ref{fig:real_data_scatter_plot} shows a nonlinear relation between the two components. Furthermore, the marginal density plots (Figure \ref{fig:real_data(density_plot)}) of the individual components shows that the data has heavy tails implying non-Gaussianity. Hence, the explanatory analysis provides evidence towards a multivariate non-stationary covariance and non-Gaussian distribution. Traditional Kriging methodologies cannot accomodate these features.

\begin{figure}[h!]
    \centering
    \includegraphics[scale=0.2]{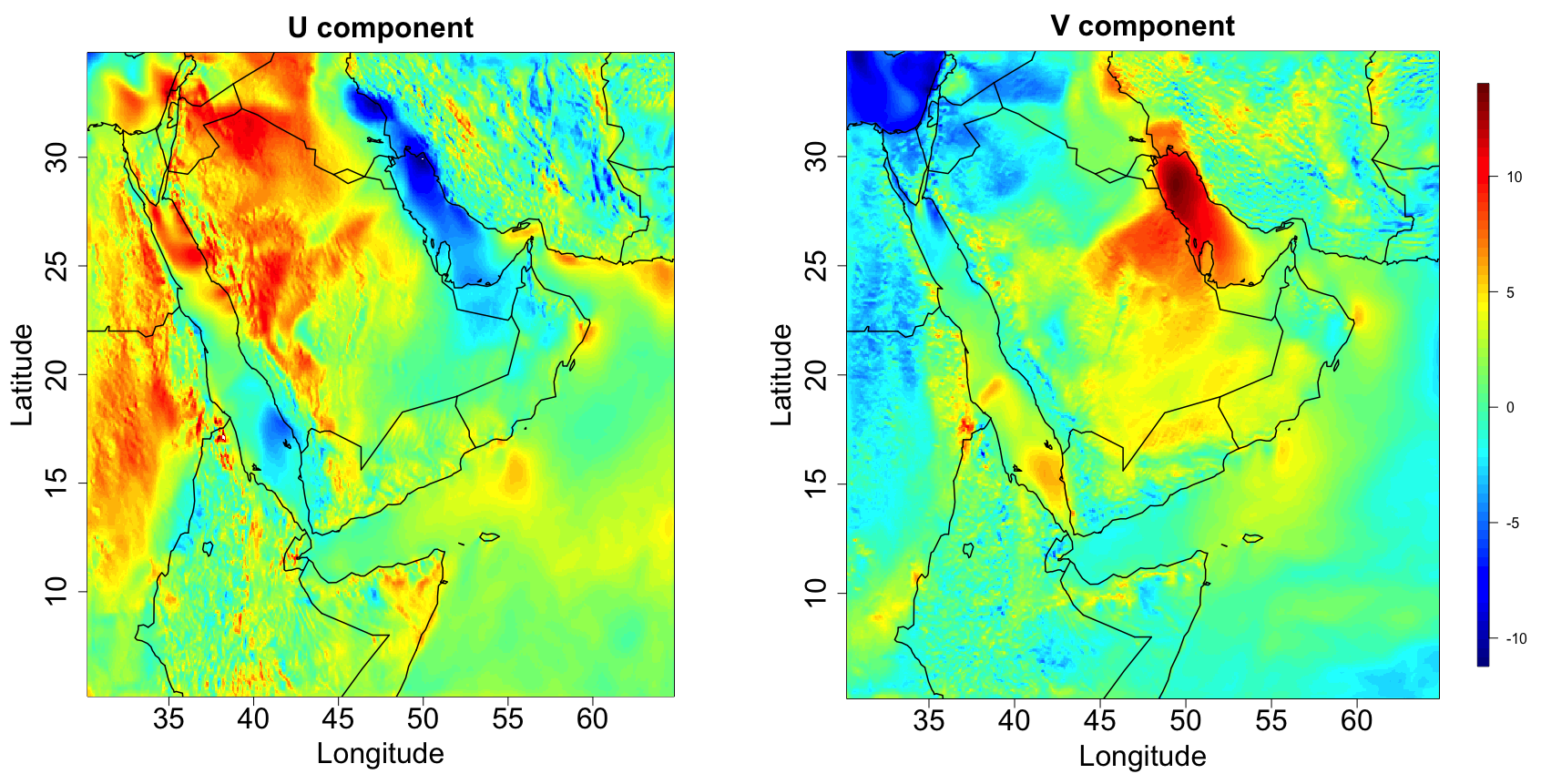} 
\caption{Visualization of the $U$ and $V$ components of the wind field over the Middle East region}
    \label{fig:12}
\end{figure}

\begin{figure}[h!]
    \centering
    \includegraphics[scale=0.3]{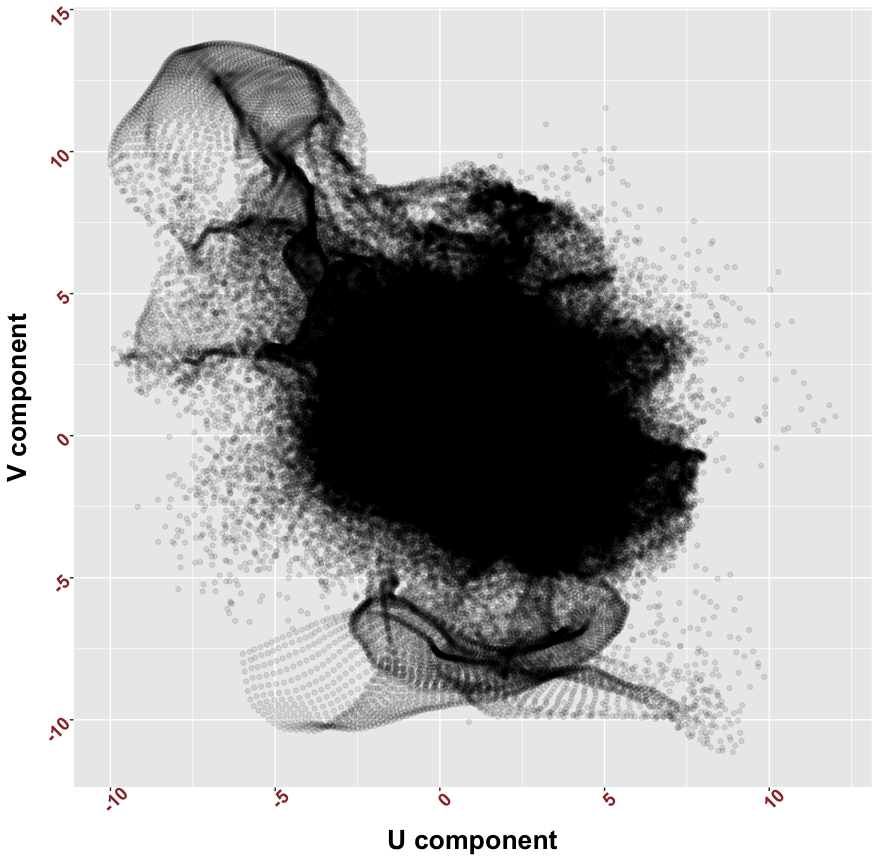} 
\caption{Scatter plot of the $U$ and $V$ components of the Middle Eastern wind field}
    \label{fig:real_data_scatter_plot}
\end{figure}

\begin{figure}[h!]
    \centering
    \includegraphics[scale=0.27]{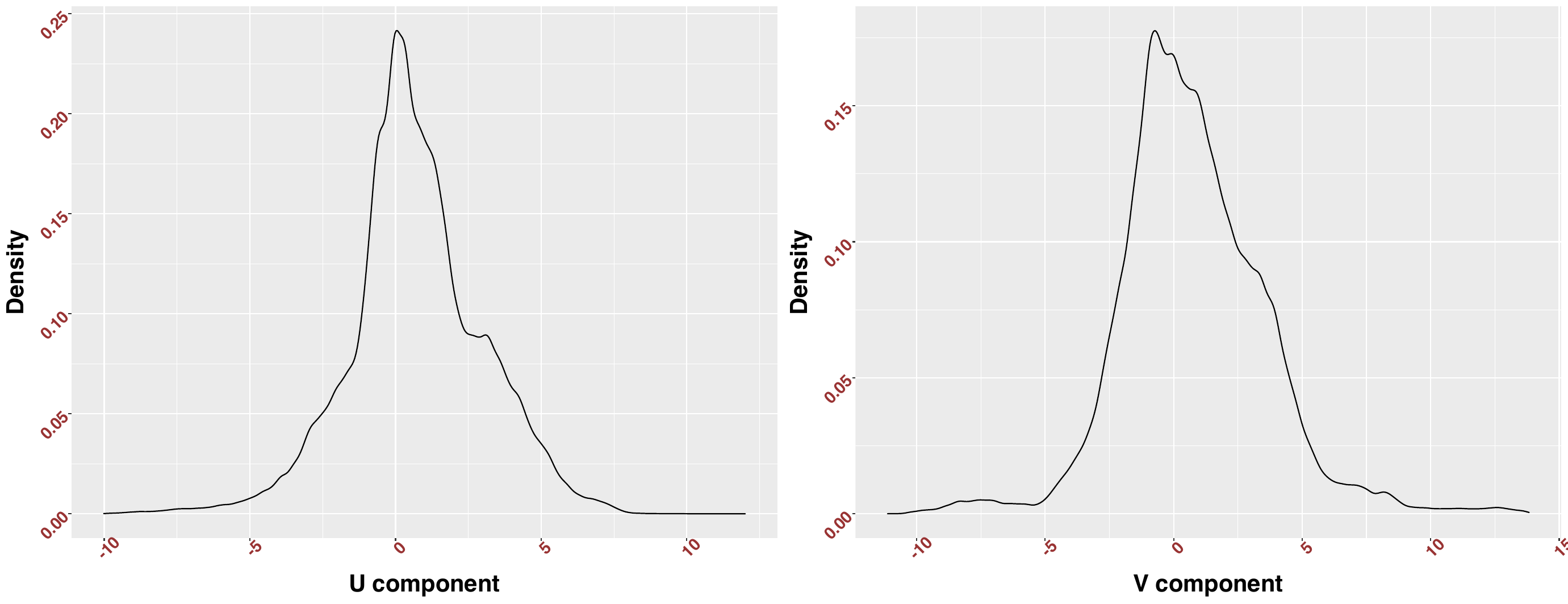} 
\caption{Sample marginal density plots of the $U$ and $V$ components of the Middle Eastern wind data}
    \label{fig:real_data(density_plot)}
\end{figure}

\newpage
\item[Simulations details and additional results:]

In this section, we provide details regarding comparisons involving Gaussian Processes (GP). We generate two-dimensional stationary Gaussian processes utilizing the parsimonious Matérn covariance function. The simulated data is derived from a zero-mean GP with bivariate Matérn covariance characterized by a cross-correlation coefficient of $\rho=0.8$, variances $\sigma_1 = 0.89$ and $\sigma_2 = 1.3$, smoothness parameters $\nu_1 = \nu_2 = 0.8$, and range parameters $\alpha_1 = 0.2$ and $\alpha_2 = 0.4$. 

We generate $100$ replicates employing the aforementioned parameter configuration on $1200$ irregularly spaced locations spanning $[0,1] \times [0,1]$. For training purposes, we utilize 1080 points, reserving the remainder for testing. The results can be seen in Table \ref{tab:2}.

Although the nonlinear structure of BDK captures more specific features from data we still need to validate that our proposal for extending the univariate DeepKriging \citep{chen2020deepkriging} to Bivariate scenario improves the prediction results by exploiting the dependence within the variables. To do this, we compare our proposed model with the univariate DeepKriging model predictions. We compute the average {MSPE} independently for each variable over $100$ replicates. Table \ref{tab:2} shows that the Bivariate DeepKriging provides better performance in the prediction error than the independent univariate DeepKriging ({IndepDK}) fitted with the same network architecture. 

\begin{table}[!htb]
    \centering
    \caption{Comparison of {MSPE} for both the variables and its standard error ({SE}) for stationary Gaussian processes.}
    \resizebox{\columnwidth}{!}{%
        \begin{tabular}{||c c c c c c||} 
        \hline
        Simulation type & Models & ${MSPE}_1$ & ${SE}_1$  & ${MSPE}_2$
        &${SE}_2$ \\ [0.5ex] 
         \hline\hline
         Gaussian & 
         \begin{tabular}{c} $CMK_{true}$\\
         {IndepDK} \\
         BDK\\ \end{tabular} & 
         \begin{tabular}{c}
         \textbf{0.23}\\ 
         0.25\\
         0.24\\
         \end{tabular} &
         \begin{tabular}{c}
         0.91 ($\times 10^{-4}$)\\ 
         1.21 ($\times 10^{-4}$)\\
         1.33 ($\times 10^{-4}$)\\
         \end{tabular} &
         \begin{tabular}{c} 
         \textbf{0.21}\\ 
         0.26\\
         0.24 \\
         \end{tabular} &
         \begin{tabular}{c} 
         0.93 ($\times 10^{-4}$)\\ 
         2.32 ($\times 10^{-4}$)\\
         1.17 ($\times 10^{-4}$)\\
         \end{tabular} \\
         \hline
    \end{tabular}%
    }
    \label{tab:2}
\end{table}

\item[Nonstationary simulation:]
The simulation focuses on bivariate data with non-stationary feature. To do so, we followed several computer examples \citep{ba2012composite,xiong2007non} to generate the non-stationary data by deterministic functions which gives non-stationarity in the data. We simulate $\mathbf{Y(s)}$ with $\boldsymbol{\mu}(\bfs) = \{ \mu_1(\bfs),\mu_2(\bfs)\} $ where 
$$
\begin{aligned}
    \mu_1(\bfs) &=  \sin{\{5(\Bar{s} - 0.9)\}}\cos{\{25(\Bar{s} - 0.9)^4\}} + \frac{(\Bar{s} - 0.9)}{2} \\
    \mu_2(\bfs) &= \sin{\{2(\Bar{s} - 0.9)\}}\cos{\{30(\Bar{s} - 0.9)^4\}} - \frac{(\Bar{s} - 0.9)}{2}
\end{aligned}
$$
$\Bar{s} = \frac{x + y}{2}, \text{where} \ \bfs = \{x,y\}$
and $\boldsymbol{\gamma}(\bfs)$ follows bivariate $\textbf{GP}$ with variance variances $ \sigma_1  = 0.01 \ \text{and} \ \sigma_2 = 0.01$ and the remaining covariance parameters are same as the Gaussian simulation parameterization . The graphical plot of the data can be viewed in Figure \ref{fig:nonstat_data}.

\begin{figure}[h]
    \centering
   \includegraphics[scale=0.35]{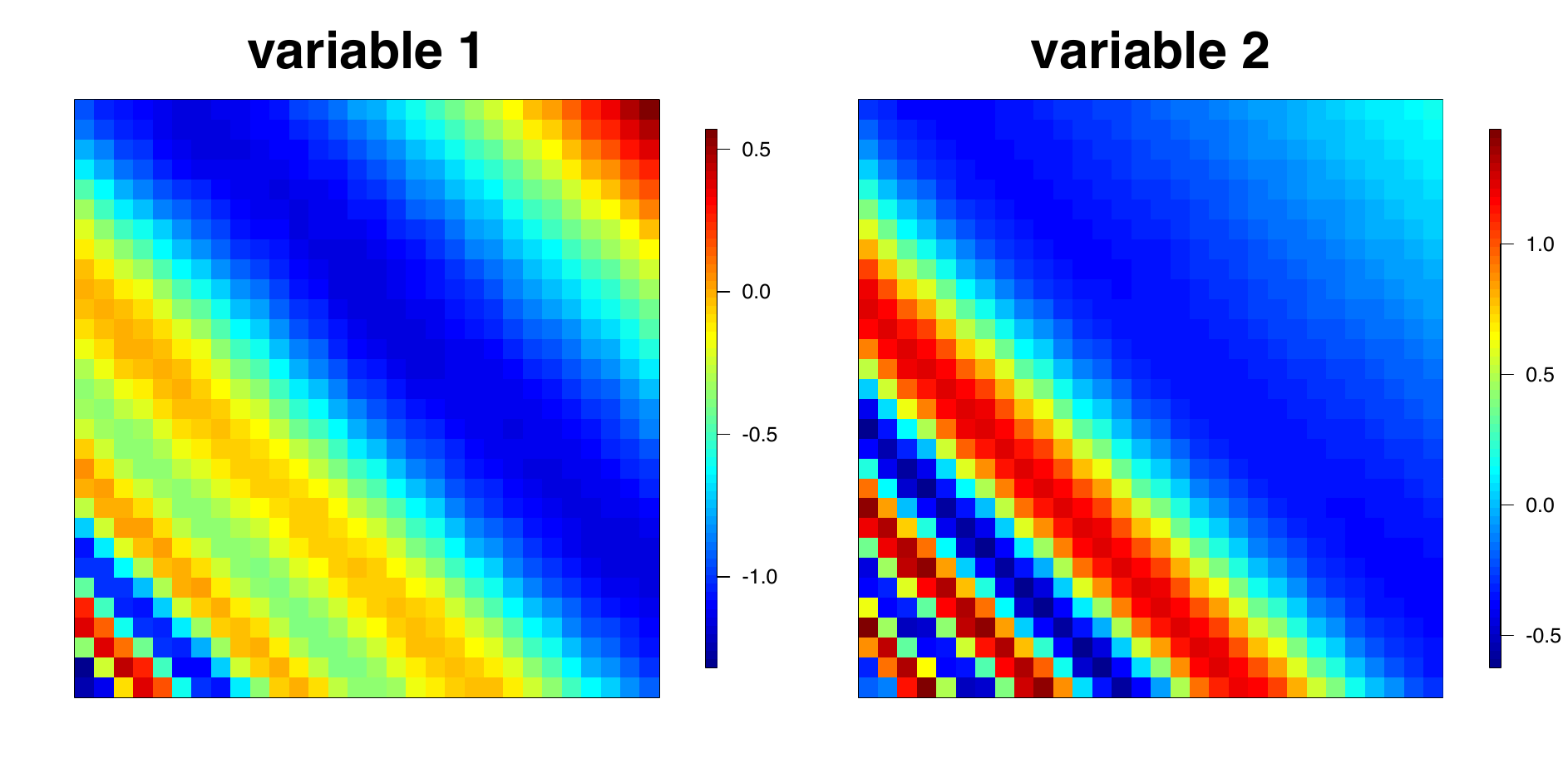}
   \caption{Visualization of the simulated nonstationary data.}
    \label{fig:nonstat_data}
\end{figure}

\begin{figure}[h]
    \centering
   \includegraphics[scale=0.36]{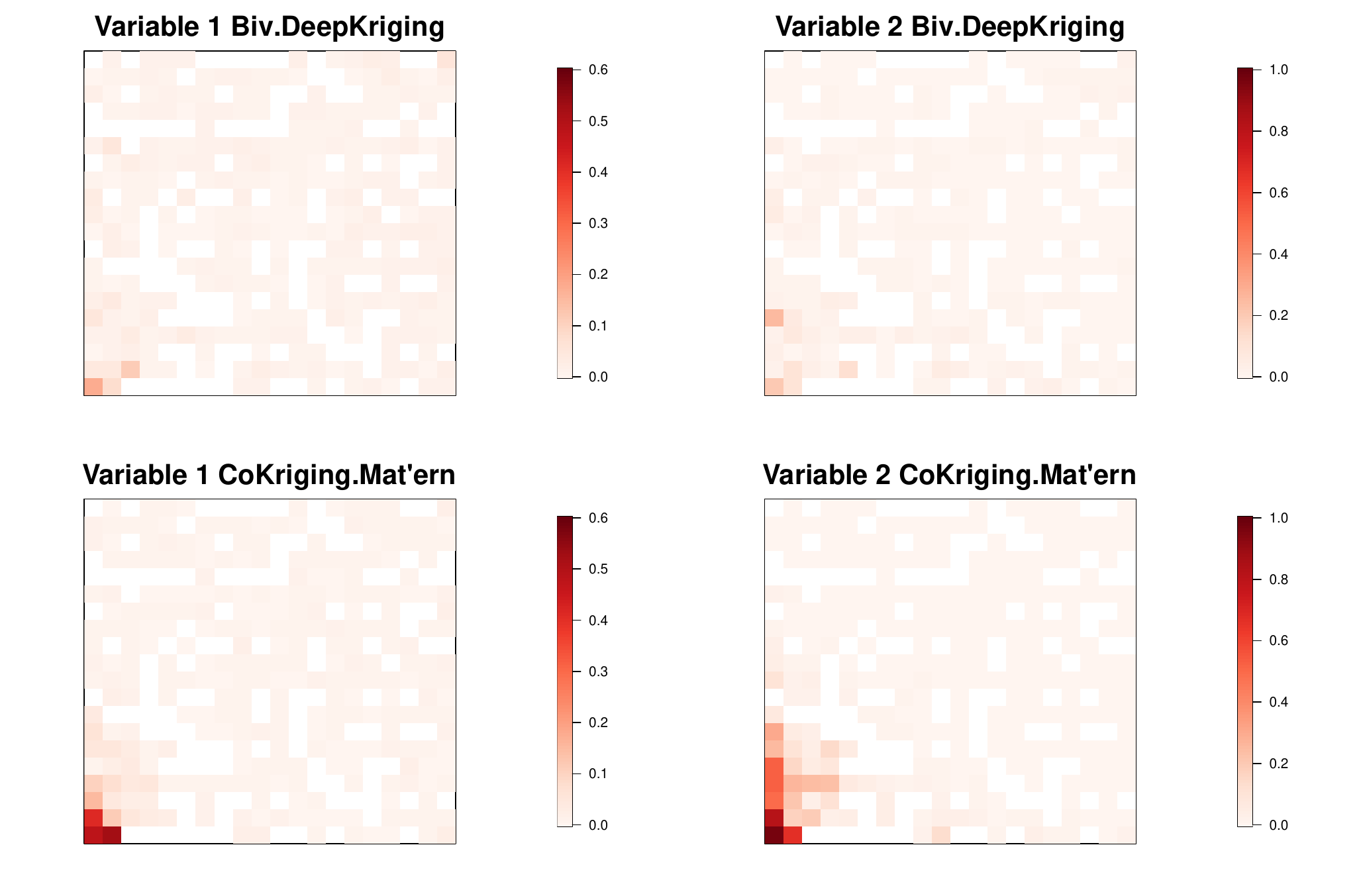}
   \caption{Comparison of absolute deviation of predictions from the truth between BDK and CMK. We computed these deviations for the test locations for both the variables.}
    \label{fig:6}
\end{figure}

From Figure \ref{fig:nonstat_data}, it is evident that the dataset exhibits roughness in the lower-left region, while appearing comparatively smoother towards the top. This observation suggests that employing the same model structure across the entire region may not be appropriate. The CoKriging model structure overlooks this spatial nonstationarity, whereas bivariate DeepKriging demonstrates adaptability to such variations, emerging as a more flexible modeling option. To further substantiate this claim, we computed the absolute error deviation for each test location individually and compared the results with the CMK model, which is the second best-performing model after BDK. In Figure \ref{fig:6}, it can be observed that the absolute error remains minimal for both models over the smoother surface. However, towards the lower-left region, the absolute error significantly increases for CMK compared to BDK. This discrepancy implies that the CoKriging model fails to capture the spatial nonstationarity adequately, whereas bivariate DeepKriging demonstrates superior performance in this aspect.

\item[Details on Neural Network Training:]

We employed the following deep neural network architecture for modeling:

\begin{itemize}
    \item We utilized $3$ layers of basis functions with $K = 4, 16, 25$ respectively, resulting in a total of $45$ basis functions for all our simulation scenarios.
    \item The neural network comprises $8$ layers with $M_1 = M_2 = M_3 = M_4 = M_5 = 100$ for the first $5$ layers, and $M_6 = M_7 = 50$, $M_8 = 2$ nodes in the final three layers.
    \item Weight matrices were initialized by sampling from a uniform distribution.
    \item The training dataset was split into a 90\%-10\% ratio for training and validation on each epoch, and the model showing the best performance on validation was saved.
    \item Hyperparameters such as learning rate \citep{li2019towards}, number of epochs, and batch size were chosen based on validation. A step-by-step approach to choosing these hyperparameters is explained below:
    \begin{itemize}
        \item Initially, a certain number of layers $L$ and nodes per layer $M$ were chosen based on prior experience, considering the nature of the dataset, which is spatial data in this case, and building upon previous work \cite{chen2020deepkriging}.
        \item Once $L$ and $M$ were determined, attention shifted to other hyperparameters.
        \item A high value for the number of epochs was set initially, and the learning rate was adjusted. If optimization did not occur due to a high learning rate causing the training loss to plateau, the learning rate was decreased gradually.
        \item After fixing the learning rate, different batch sizes were tested to observe their effect on the training loss.
        \item Based on observations, the number of epochs required for complete optimization became apparent.
        \item If results were unsatisfactory, $L$ and $M$ were adjusted iteratively. Overfitting was monitored by examining the training and validation loss.
    \end{itemize}
    \item Through iterative experiments, optimal hyperparameters were selected. For non-Gaussian simulations, we used a learning rate of $0.001$, batch size of $512$, and a total of $1250$ epochs. For nonstationary experiments, a learning rate of $0.001$, batch size of $256$, and $1500$ epochs were chosen.
\end{itemize}

\item[Performance of BDK conditioned on number of layers:]

We conducted a comparative study involving three different DNN
architectures for both non-Gaussian simulations and real data. Table \ref{tab:layer_based_comparison} gives a comprehensive overview on the performance of BDK in varied architectures. We denoted the number of layers for each model by the subscript added to it. It can be seen from the MSPEs that $BDK_{L=2}$ underfits the data, $BDK_{L=15}$ overfits it. The optimum number of layers $L=8$ provides best result here. 
\begin{table}[!htb]
    \centering
    \caption{Comparison of average {MSPE} for both the variables for different scenarios.}
    {%
        \begin{tabular}{||c c c c||} 
        \hline
        data type & Model & ${MSPE}_1$ & ${MSPE}_2$
        \\ [0.5ex] 
         \hline\hline
         non-Gaussian & 
         \begin{tabular}{c} $BDK_{L=2}$\\
         $BDK_{L=8}$ \\
         $BDK_{L=15}$\\ \end{tabular} & 
         \begin{tabular}{c}
         35.94\\ 
         32.71\\
         32.88\\
         \end{tabular} &
         \begin{tabular}{c} 
         48.70 \\ 
         23.83 \\
         24.33 \\
         \end{tabular} \\
         \hline
         real data & 
         \begin{tabular}{c} $BDK_{L=2}$\\
         $BDK_{L=8}$ \\
         $BDK_{L=15}$\\ \end{tabular} & 
         \begin{tabular}{c}
         0.732\\ 
         0.394\\
         0.433\\
         \end{tabular} &
         \begin{tabular}{c} 
         0.875 \\ 
         0.392 \\
         0.413 \\
         \end{tabular} \\
         \hline
    \end{tabular}%
    }
    \label{tab:layer_based_comparison}
\end{table}

\item[Prediction interval computation:] 

Below we give necessary details on the prediction interval calculation as implemented in the code. 

\begin{enumerate}
\item We import the necessary training and testing csv files. 
    \begin{itemize}  
        \item Spatial coordinates (\texttt{s\_train} and \texttt{s\_test}) are extracted, and response variables are standardized using mean and variance normalization.
        \item Covariates are scaled using Min-Max normalization for better model training.
    \end{itemize}
    \item Basis functions are constructed for spatial data using Wendland kernels, enabling spatial representation in higher dimensions.
    \item Separate basis functions are computed for training and testing data.
    \item Data splitting is performed for ensemble training and Mean Squared Error (MSE) evaluation. We first split the data into training and testing. Then training data is further decomposed into \texttt{train\_base}, \texttt{train\_ensamble} and \texttt{train\_mse}. Note that,  \texttt{train\_base} $\in$ \texttt{train\_ensamble}.
    \item A feedforward neural network is defined using Keras, with:
    \begin{itemize}
        \item Five dense layers with ReLU activation.
        \item A final dense layer with linear activation for output.
    \end{itemize}
    \item The base model is compiled with the Adam optimizer and trained on \texttt{train\_base}.
    \item An ensemble of models is trained using the custom \texttt{fit\_ensemble} function. The \texttt{fit\_ensemble} function freezes the first $L_0 = 3$ layers of the model and just trains the last two layers for each bootstrap sample from \texttt{train\_ensamble}. Note that, the choice of $L_0$ depends on model calibration. We found the number of layers $L_0$ that matches closely to the Gaussian Kriging results and then used the same number for other simulation and the real data scenario.
    \item Prediction intervals are computed:
    \begin{itemize}
        \item Mean and variance are predicted for the training and testing data using the \texttt{train\_ensamble} for the test locations. 
        \item Random error components are calculated using the nearest neighborhood approach with \texttt{train\_mse}.
        \item Upper and lower bounds for prediction intervals are computed using a 95\% confidence level with normal quantiles. 
    \end{itemize}
    \item The Prediction Interval Coverage Probability (PICP) and Mean Prediction Interval Width (MPIW) are computed for each simulation.
\end{enumerate}

\newpage
\item[The Shapley value:] 

The Shapley value of a feature value represents its contribution to the payout, weighted and summed over all possible feature value combinations:
\[
\psi_j(\mathrm{val}) = \sum_{S \subseteq \{1, \ldots, p\} \backslash \{j\}} \frac{|S| !(p - |S| - 1)!}{p!} (\operatorname{val}(S \cup \{j\}) - \operatorname{val}(S))
\]
where $S$ is a subset of the features used in the model, $\mathbf{x}$ is the vector of feature values of the instance to be explained, and $p$ is the number of features. $\operatorname{val}_x(S)$ is the prediction for feature values in set $S$ that are marginalized over features not included in set $S$. From this point onward, we denote $\hat{f}_{NN_1}(\cdot)$ as $\hat{f}(\cdot)$:
\[
\operatorname{val}_x(S) = \int \hat{f}(x_1, \ldots, x_p) \, d\mathbb{P}_{x \notin S} - E_X(\hat{f}(X))
\]
where $X$ is a random variable of the feature.

Multiple integrations are performed for each feature not contained in $S$. For example, for a model with five features $x_1, x_2, x_3, x_4$, and $x_5$, and evaluating the prediction for the coalition $S$ consisting of feature values $x_1$ and $x_3$:
\[
\operatorname{val}_x(S) = \operatorname{val}_x(\{1,3\}) = \int_{\mathbb{R}} \int_{\mathbb{R}} \hat{f}(x_1, X_2, x_3, X_4, X_5) \, d\mathbb{P}_{X_2 X_4X_5} - E_X(\hat{f}(X))
\]

The Shapley value is the only attribution method that satisfies the properties of Efficiency, Symmetry, Dummy, and Additivity, which together can be considered a definition of a fair payout.

\textbf{Estimating the Shapley Value}

To calculate the exact Shapley value, all possible coalitions (sets) of feature values must be evaluated with and without the $j$-th feature. However, for a large number of features, the exact solution becomes problematic due to the exponentially increasing number of possible coalitions. An approximation with Monte Carlo sampling can be written as:
\[
\hat{\psi}_j = \frac{1}{M} \sum_{m=1}^M (\hat{f}(x_{+j}^m) - \hat{f}(x_{-j}^m))
\]
where $\hat{f}(x_{+j}^m)$ is the prediction for $\mathbf{x}$ with a random number of feature values replaced by feature values from a random data point $\mathbf{z}$, except for the respective value of feature $j$. The $\mathbf{x}$-vector $x_{-j}^m$ is almost identical to $x_{+j}^m$, except the value $x_j^m$ is also taken from the sampled $\mathbf{z}$. Each of these $M$ new instances is a combination of two instances, akin to "Frankenstein's Monster." 

The following algorithm outlines the approximate Shapley estimation for a single feature value:

\textbf{Algorithm:} Approximate Shapley estimation for a single feature value
\begin{itemize}
    \item \textbf{Output:} Shapley value for the value of the $j$-th feature
    \item \textbf{Required:} Number of iterations $M$, instance of interest $\mathbf{x}$, feature index $j$, data matrix $\mathbf{X}$, and machine learning model $f$
    \item For all $m = 1, \ldots, M$:
    \begin{itemize}
        \item Draw random instance $\mathbf{z}$ from the data matrix $\mathbf{X}$
        \item Choose a random permutation $o$ of the feature values
        \item Order instance $\mathbf{x}$: $\mathbf{x}_o = (x_{(1)}, \ldots, x_{(j)}, \ldots, x_{(p)})$
        \item Order instance $\mathbf{z}$: $\mathbf{z}_o = (z_{(1)}, \ldots, z_{(j)}, \ldots, z_{(p)})$
        \item Construct two new instances:
        \begin{itemize}
            \item With $j$: $\mathbf{x}_{+j} = (x_{(1)}, \ldots, x_{(j-1)}, x_{(j)}, z_{(j+1)}, \ldots, z_{(p)})$
            \item Without $j$: $\mathbf{x}_{-j} = (x_{(1)}, \ldots, x_{(j-1)}, z_{(j)}, z_{(j+1)}, \ldots, z_{(p)})$
        \end{itemize}
        \item Compute marginal contribution: $\psi_j^m = \hat{f}(\mathbf{x}_{+j}) - \hat{f}(\mathbf{x}_{-j})$
    \end{itemize}
    \item Compute Shapley value as the average: $\psi_j(\mathbf{x}) = \frac{1}{M} \sum_{m=1}^M \psi_j^m$
\end{itemize}

The procedure must be repeated for each feature to obtain all Shapley values.

We utilize the Shapley value method to interpret the covariates associated with the non-Gaussian scenario. Several examples are provided below. Figure \ref{fig:cov_dep} illustrates the partial dependence of covariate 5 on variable 1, while Figure \ref{fig:cov_imp} demonstrates the contribution of each covariate on variable 1 for the test location $\mathbf{s}_{35}$. 

% The code on Shapley values is available in the following link \url{https://github.com/pratik187/Bivariate_DeepKriging/blob/main/python_scripts/ipynb_files/shapley_values.ipynb}. 

\begin{figure}[h]
    \centering
   \includegraphics[scale=0.45]{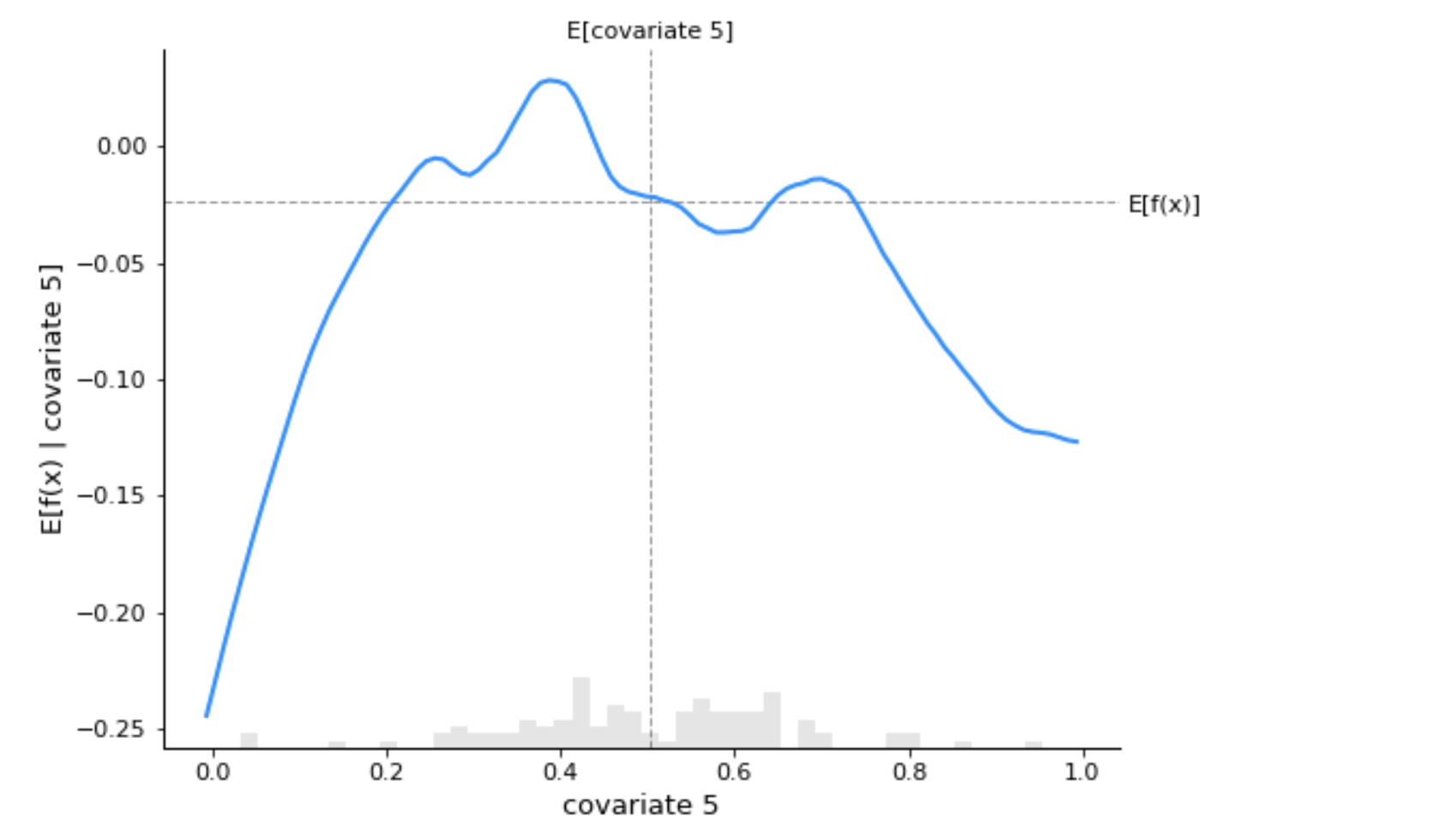}
   \caption{Visualization of the dependence of Covariate 5 on the response.}
    \label{fig:cov_dep}
\end{figure}

\begin{figure}[h]
    \centering
   \includegraphics[scale=0.45]{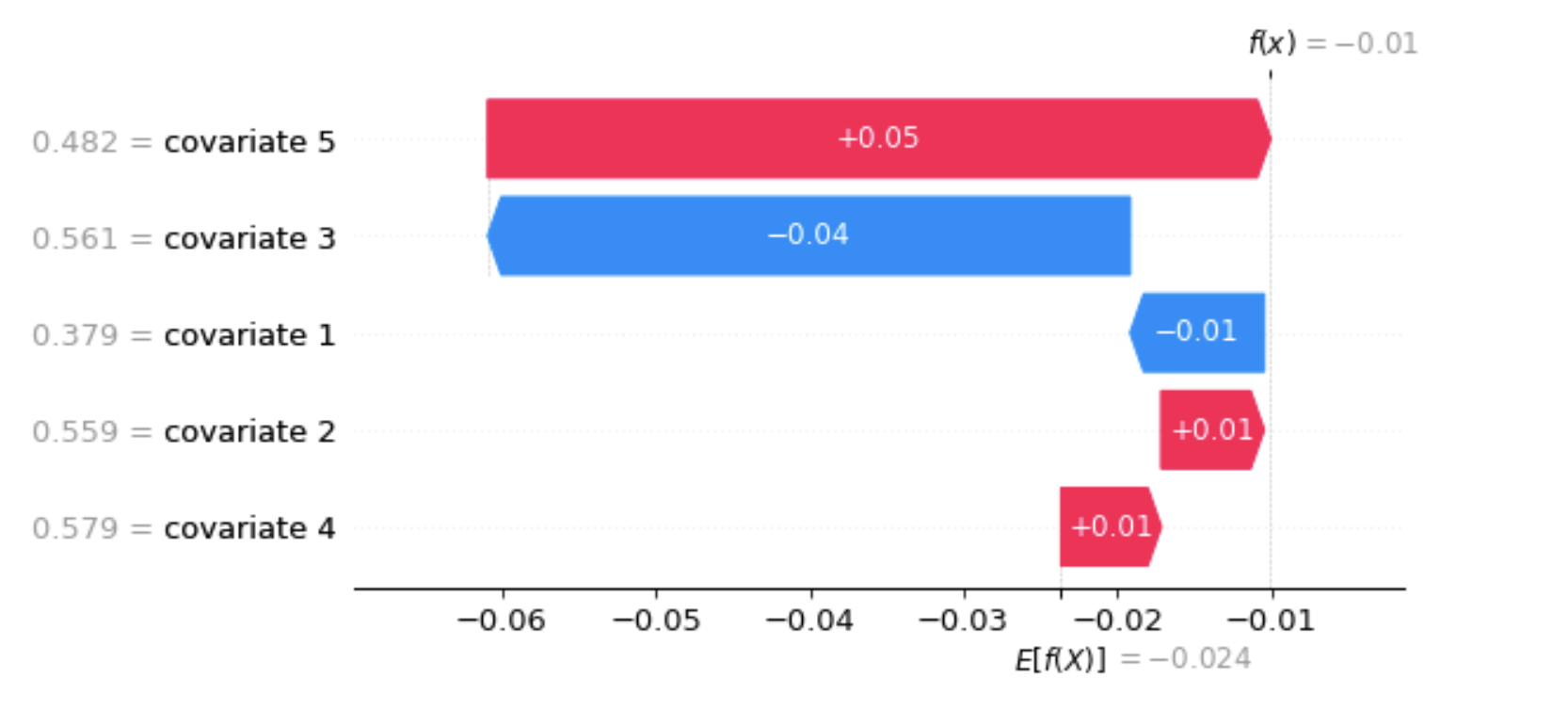}
   \caption{A comparative study of all covariate contributions on the response at the test location $\bfs_{35}$.}
    \label{fig:cov_imp}
\end{figure}
    
    % For more information, see this article on \href{https://towardsdatascience.com/optimizing-deep-neural-networks-through-hyperparameter-tuning-1c8ae15bd3c7}{Hyper-parameter tuning}.

\end{description}

\newpage

\bibliographystyle{chicago}

\bibliography{Bibliography-MM-MC}